%% file: arxiv_main.tex
\documentclass{article}

\usepackage{microtype}
\usepackage{graphicx}
\usepackage{booktabs} 
\usepackage{xcolor}
\usepackage{subcaption}
\usepackage{hyperref}
\definecolor{mydarkblue}{rgb}{0,0.08,0.45}
  \hypersetup{ %
    pdftitle={},
    pdfsubject={},
    pdfkeywords={},
    pdfborder=0 0 0,
    pdfpagemode=UseNone,
    colorlinks=true,
    linkcolor=mydarkblue,
    citecolor=mydarkblue,
    filecolor=mydarkblue,
    urlcolor=mydarkblue,
    }

\usepackage{natbib}
\renewcommand{\cite}[1]{\citep{#1}}
\setcitestyle{authoryear,round,citesep={;},aysep={,},yysep={;}}
\usepackage{arxiv}

\usepackage{amsmath}
\usepackage{amssymb}
\usepackage{mathtools}
\usepackage{amsthm}
\usepackage{thm-restate}
\input{macros}

\title{Linear Contextual Bandits with Hybrid Payoff: Revisited}

\renewcommand{\shorttitle}{Hybrid Linear Contextual Bandits Revisited}
\renewcommand{\headeright}{}
\renewcommand{\undertitle}{}

\author{ 
	Nirjhar Das \thanks{Microsoft Research, 
	Bengaluru, India. \texttt{\href{mailto:nirjhar.das@alumni.iitd.ac.in}{nirjhar.das@alumni.iitd.ac.in}}} 
    \And
    Gaurav Sinha \thanks{Microsoft Research, 
	Bengaluru, India.  \texttt{\href{mailto:gauravsinha@microsoft.com}{gauravsinha@microsoft.com}}} 
}

\date{}

\begin{document}

\maketitle

\begin{abstract}
    We study the Linear Contextual Bandit (\lincb) problem in the hybrid reward setting. In this setting, every arm's reward model contains arm specific parameters in addition to parameters shared across the reward models of all the arms. We can easily reduce this setting to two closely related settings; (a) \emph{Shared} - no arm specific parameters, and (b) \emph{Disjoint} - only arm specific parameters, enabling the application of two popular state of the art algorithms - \LinUCB\ and \DisLinUCB (proposed as Algorithm $1$ in Li et al. 2010). When the arm features are stochastic and satisfy a popular diversity condition, we provide new regret analyses for both \LinUCB\ and \DisLinUCB\ that significantly improves upon the known regret guarantees of these algorithms. Our novel analysis critically exploits the structure of the hybrid rewards and diversity of the arm features. Along with proving these new guarantees, we  introduce a new algorithm \HyLinUCB\ that crucially modifies \LinUCB\ (using a new exploration coefficient) to account for sparsity in the hybrid setting. Under the same diversity assumptions, we prove that at the end of $T$ rounds, \HyLinUCB\ also incurs only $\tilde{O}(\sqrt{T})$ regret. We perform extensive experiments on synthetic and real-world datasets demonstrating strong empirical performance of \HyLinUCB. When the number of arm specific parameters is much larger than the number of shared parameters, we observe that \DisLinUCB\ incurs the lowest regret. In this case, regret of \HyLinUCB\ is the second best and it is extremely competitive to \DisLinUCB. In all other situations, including our real-world dataset, \HyLinUCB\ has significantly lower regret than \LinUCB, \DisLinUCB\ and other state of the art baselines we considered. We also empirically observe that the regret of \HyLinUCB\ grows much slower with the number of arms $K$, compared to baselines, making it suitable even for very large action spaces.
\end{abstract}

\keywords{Linear Contextual Bandits, Hybrid Payoff, Stochastic Contexts}

\input{introduction}

\input{formulation}
\input{algorithms}
\input{datasets}
\input{conclusion}

\bibliographystyle{plainnat}
\bibliography{references}

\newpage
\appendix
\appendixtitle

\input{regret_linucb_appendix}
\input{regret_hylinucb_appendix}
\input{regret_dislinucb_appendix}
\input{concentration_appendix}
\input{lin_alg_appendix}
\input{empirical-validation-assumption}

\end{document}

%% file: macros.tex
\usepackage{xspace}
\usepackage{amsthm}
\usepackage{amsmath, amsfonts, amssymb}
\usepackage{mathtools}
\usepackage{bm, url}

\newcommand{\lincb}{\texttt{LinearCB}\xspace}

\newcommand{\HyLinUCB}{\texttt{HyLinUCB}\xspace}
\newcommand{\DisLinUCB}{\texttt{DisLinUCB}\xspace}

\newcommand{\SupLinUCB}{\texttt{SupLinUCB}\xspace}

\newcommand{\HyLinCB}{\texttt{HyLinCB}\xspace}
\newcommand{\LinCB}{\texttt{LinCB}\xspace}
\newcommand{\LinUCB}{\texttt{LinUCB}\xspace}
\newcommand{\HyRan}{\texttt{HyRan}\xspace}

\newcommand{\Reg}{\texttt{Reg}}

\newcommand{\alglinelabel}{%
  \addtocounter{ALC@line}{-1}
  \refstepcounter{ALC@line}
  \label
}

\newcommand{\Ab}{\mathbf{A}}
\newcommand{\Bb}{\mathbf{B}}
\newcommand{\Cb}{\mathbf{C}}
\newcommand{\Db}{\mathbf{D}}

\newcommand{\Hb}{\mathbf{H}}
\newcommand{\Ib}{\mathbf{I}}

\newcommand{\Mb}{\mathbf{M}}

\newcommand{\Ub}{\mathbf{U}}
\newcommand{\Vb}{\mathbf{V}}
\newcommand{\Wb}{\mathbf{W}}
\newcommand{\Xb}{\mathbf{X}}

\newcommand{\Zb}{\mathbf{Z}}
\newcommand{\R}{\mathbb{R}}
\newcommand{\N}{\mathbb{N}}

\newcommand{\cE}{\mathcal{E}}
\newcommand{\cF}{\mathcal{F}}

\newcommand{\cH}{\mathcal{H}}

\newcommand{\cN}{\mathcal{N}}

\newcommand{\cP}{\mathcal{P}}

\newcommand{\cT}{{\mathcal{T}}}

\newcommand{\cX}{\mathcal{X}}

\newcommand{\EE}{\mathbb{E}} 

\newcommand{\ind}[1]{\mathbb{I}\cbr{#1}} 
\newcommand{\NN}{\mathbb{N}} 
\newcommand{\PP}{\mathbb{P}} 
\newcommand{\RR}{\mathbb{R}} 
\newcommand*{\argmin}{\mathop{\mathrm{argmin}}}
\newcommand*{\argmax}{\mathop{\mathrm{argmax}}}

\newcommand*{\zero}{{\mathbf{0}}}

\newcommand{\diag}{{\rm diag}}

\newcommand{\thta}{\theta^*}
\newcommand{\thth}{\widehat{\theta}}
\newcommand{\bta}[1]{\beta^*_{#1}}
\newcommand{\bth}[1]{\widehat{\beta}_{#1}}
\newcommand{\thtill}{\widehat{\phi}}
\newcommand{\thbar}{{\phi^*}}
\newcommand{\phist}[1]{\phi^*_{#1}}
\newcommand{\phiht}[1]{\widehat{\phi}_{#1}}

\newcommand{\norm}[2]{\left\lVert#1\right\rVert_{#2}}

\newcommand{\dotp}[2]{\langle{#1},{#2}\rangle}

\newcommand{\rbr}[1]{\left(#1\right)}
\newcommand{\sbr}[1]{\left[#1\right]}
\newcommand{\cbr}[1]{\left\{#1\right\}}

\newcommand{\abr}[1]{\left|#1\right|}

\ifx\BlackBox\undefined
\newcommand{\BlackBox}{\rule{1.5ex}{1.5ex}}  
\fi

\ifx\QED\undefined
\def\QED{~\rule[-1pt]{5pt}{5pt}\par\medskip}
\fi

\ifx\proof\undefined
\newenvironment{proof}{\par\noindent{\bf Proof\ }}{\hfill\BlackBox\\[2mm]}
\fi

\ifx\theorem\undefined
\newtheorem{theorem}{Theorem}
\fi
\ifx\example\undefined
\newtheorem{example}{Example}
\fi
\ifx\property\undefined
\newtheorem{property}{Property}
\fi
\ifx\lemma\undefined
\newtheorem{lemma}[theorem]{Lemma}
\fi
\ifx\proposition\undefined
\newtheorem{proposition}[theorem]{Proposition}
\fi
\ifx\remark\undefined
\newtheorem{remark}[theorem]{Remark}
\fi
\ifx\corollary\undefined
\newtheorem{corollary}[theorem]{Corollary}
\fi
\ifx\definition\undefined
\newtheorem{definition}[theorem]{Definition}
\fi
\ifx\conjecture\undefined
\newtheorem{conjecture}[theorem]{Conjecture}
\fi
\ifx\fact\undefined
\newtheorem{fact}[theorem]{Fact}
\fi
\ifx\claim\undefined
\newtheorem{claim}[theorem]{Claim}
\fi
\ifx\assum\undefined
\newtheorem{assum}[theorem]{Assumption}
\fi
\ifx\exercise\undefined
\newtheorem{exercise}{Exercise}
\fi
\ifx\assumption\undefined
\newtheorem{assumption}[theorem]{Assumption}
\fi

\usepackage{prettyref}

\newcommand{\savehyperref}[2]{\texorpdfstring{\hyperref[#1]{#2}}{#2}}
\newrefformat{eq}{\savehyperref{#1}{\textup{(\ref*{#1})}}}
\newrefformat{eqn}{\savehyperref{#1}{Equation~\ref*{#1}}}
\newrefformat{lem}{\savehyperref{#1}{Lemma~\ref*{#1}}}
\newrefformat{def}{\savehyperref{#1}{Definition~\ref*{#1}}}
\newrefformat{line}{\savehyperref{#1}{line~\ref*{#1}}}
\newrefformat{thm}{\savehyperref{#1}{Theorem~\ref*{#1}}}
\newrefformat{corr}{\savehyperref{#1}{Corollary~\ref*{#1}}}
\newrefformat{sec}{\savehyperref{#1}{Section~\ref*{#1}}}
\newrefformat{ssec}{\savehyperref{#1}{Subsection~\ref*{#1}}}
\newrefformat{app}{\savehyperref{#1}{Appendix~\ref*{#1}}}
\newrefformat{ass}{\savehyperref{#1}{Assumption~\ref*{#1}}}
\newrefformat{ex}{\savehyperref{#1}{Example~\ref*{#1}}}
\newrefformat{fig}{\savehyperref{#1}{Figure~\ref*{#1}}}
\newrefformat{alg}{\savehyperref{#1}{Algorithm~\ref*{#1}}}
\newrefformat{rem}{\savehyperref{#1}{Remark~\ref*{#1}}}
\newrefformat{conj}{\savehyperref{#1}{Conjecture~\ref*{#1}}}
\newrefformat{prop}{\savehyperref{#1}{Proposition~\ref*{#1}}}
\newrefformat{proto}{\savehyperref{#1}{Protocol~\ref*{#1}}}
\newrefformat{prob}{\savehyperref{#1}{Problem~\ref*{#1}}}
\newrefformat{claim}{\savehyperref{#1}{Claim~\ref*{#1}}}
\newrefformat{fact}{\savehyperref{#1}{Fact~\ref*{#1}}}

%% file: introduction.tex
\section{Introduction}
\label{section:introduction}


The \lincb problem is a popular choice to model sequential decision making scenarios such as news recommendations \citep{li2010contextual}, web-page optimization \citep{Hill2017} etc. In \lincb, at each round, a learner receives (from the environment) a set of arms (actions) specified as feature vectors, selects one of them, and receives a reward sampled from a linear model (over the features of the selected arm). The goal of the learner is to design a policy of selecting arms that minimizes its cumulative regret. Here, regret of a round is calculated as the difference between the best possible reward in that round and the actual reward received.

The \lincb\ problem is most commonly studied under the \emph{shared setting}. 
Here, the linear reward model is shared across all the arms and algorithms such as \LinUCB\ \citep{li2010contextual, lattimore2020bandit} achieve state of the art regret. However, this setting can be quite restrictive in applications where the reward model is different for different arms. For example, in news recommendation, where arms are different news items, the reward model (say click through rate) for news items about sports could be very different from that of politics, movies etc. As a result, the problem has also been studied under the \emph{disjoint} and \emph{hybrid} settings~\cite{li2010contextual}. While in the disjoint setting, the reward model of each arm only contains parameters specific to the arm (called arm-specific parameters), in the hybrid setting they also contain parameters shared across all the arms (called shared parameters). Algorithms were developed for the disjoint and hybrid settings in Algorithm $1$ and Algorithm $2$ (respectively) of \cite{li2010contextual}. While Algorithm $1$ (called \DisLinUCB\ from here on-wards) can be analyzed using ideas similar to the analysis of \LinUCB\ for the shared setting \citep{lattimore2020bandit}, Algorithm $2$ requires tuning of a hyper-parameter. Analyzing the regret of this algorithm is quite non-trivial and to the best of our knowledge tight regret guarantees for the hybrid setting are not known. Designing algorithms for the hybrid setting that overcome this challenge and also have strong regret guarantees is the main focus of our work. For each arm's reward model, let $d_1, d_2$ be the number of shared and arm specific parameters respectively, and let $K$ be the total number of arms. Further, define $d\coloneqq d_1+d_2K$. We make the following contributions.







\subsection{Our Contributions}
\label{subsection-contributions}
\begin{enumerate}
    \item First, we reduce the hybrid setting to the shared setting with $d$ parameters and provide a new analysis for \LinUCB \citep{lattimore2020bandit}. We prove that if features of arms pulled by \LinUCB\ satisfy Assumption~\ref{assumption:independent-subgaussian-features}, then, at the end of $T$ rounds, \LinUCB\ incurs a regret of $\tilde{O}(\sqrt{d K T})$, when $T = \tilde{\Omega}(K^3)$. Note that the standard guarantee of \LinUCB\ for this reduced problem is $\tilde{O}(d\sqrt{T})$.
    \item Next, by reducing to the disjoint setting, we provide a new analysis for \DisLinUCB. We prove that, under the same assumption as above, at the end of $T$ rounds, \DisLinUCB\ incurs a regret of $\tilde{O}\rbr{\sqrt{(d_1 + d_2) K T}}$, when $T = \tilde{\Omega}(1)$. The standard analysis (following analysis of \LinUCB\ in \cite{lattimore2020bandit}) implies $\tilde{O}((d_1 + d_2)\sqrt{K T})$ regret, which is much worse.
    \item Finally, we modify \LinUCB\ using a tighter exploration parameter and develop a new algorithm \HyLinUCB. Under the same diversity assumption, we prove a $\tilde{O}(\sqrt{K^3 T} + \sqrt{d K T})$ regret guarantee for this algorithm. While our guarantee has a weaker dependence on $K$ compared to the above algorithms, we empirically observe it to be much stronger compared to them. By performing extensive experiments on synthetic (capturing a wide range of problem settings) and real-world datasets (Yahoo! Front Page Dataset \cite{yahoo-dataset}), we demonstrate that in almost all cases \HyLinUCB\ has much lower regret than \LinUCB, \DisLinUCB\ and other baselines. When the number of shared parameters is larger than the arm-specific parameters, regret of \HyLinUCB\ is significantly lower than all state of the art baselines we compare it with. In the other case, i.e., when the number of arm-specific parameters are larger, \HyLinUCB\ is extremely competitive with the best algorithm i.e. \DisLinUCB. We also empirically study the variation in regret with respect to the number of arms $K$, and observe that regret of \HyLinUCB\ grows much slower with $K$, compared to baselines, making it suitable even for very large action spaces. 




\end{enumerate}

\subsection{Additional Remarks on Contributions}
\label{subsection:additional-remarks}
We make some additional remarks to provide more clarity on our contributions.

\textbf{Reduction to the shared setting:} The hybrid setting is easily reduced to the shared setting by combining the parameters of linear reward models of all the actions into a common reward model. We provide complete details of this reduction in Section \ref{section:formulation}.
This reduction helps us apply most of the known algorithms for \lincb\ (discussed in detail in Section \ref{section:related-work}). Note that the reduction significantly increases the dimensionality of the arm features and the known regret guarantees of these \lincb\ algorithms might not be optimal anymore. The additional structure in the problem needs to be exploited in the analysis to improve these guarantees. Our first contribution does this for \LinUCB. Similar improved analysis might exist for other state of the art \lincb\ algorithms we discuss in Section \ref{section:related-work}. We leave this problem for future work.


\textbf{Regret guarantee of \HyLinUCB:} Even though our regret guarantee for \HyLinUCB\ has a slightly worse dependence on $K$, we believe this is an artifact of our proof and that it can be improved further. Our synthetic and real-data experiments provide strong evidence of this as \HyLinUCB\ performs much better than \LinUCB in all the different problem settings we consider. Regret guarantee that improves this dependence is an interesting direction for future work. 

\textbf{Diversity Assumption: }
Our regret guarantees are derived under the assumption that the arms pulled by the algorithms satisfy a diversity condition (Assumption \ref{assumption:independent-subgaussian-features}) that has been studied in~\citep{chatterji2020osom, ghosh2021problem,gentile2017context}. The assumption states that the minimum eigenvalue of the expected outer product of the pulled arm's feature vector with itself, is bounded away from zero. This assumption allows the algorithms to perform parameter estimation in conjunction to regret minimization, the former being crucial to derive better dependence on the number of arms in the regret upper bound. The necessity of such diversity assumptions and their implications on better regret rates has been studied in~\citep{pmlr-v139-papini21a}. Even though the assumption is algorithm dependent, there are problem instances for which all algorithms satisfy it (Section 3, \citep{pmlr-v139-papini21a}). We empirically show that Assumption \ref{assumption:independent-subgaussian-features} is indeed satisfied by the algorithms we study in this work. 

\subsection{Related Work}
\label{section:related-work}

As highlighted in Section \ref{subsection:additional-remarks}, the \lincb\ problem in the hybrid setting can be reduced to the shared setting, enabling the application of all \lincb\ algorithms designed for the shared setting. Over the last couple of decades, there has been a substantial progress on \lincb\ for the shared setting. Many state of the art algorithms follow the optimism in the face of uncertainty approach \citep{Lai1985}, and provide regret analysis with near optimal guarantees \citep{auer2003, Dani2008, Rusmevichientong2008, chu2011contextual, abbasi2011improved}. For general stochastic arm features, the best known regret upper bound (independent of number of arms $K$) provided in these works is $\tilde{O}(d\sqrt{T})$ and the best known lower bound \citep{chu2011contextual} is $\tilde{\Omega}(d\sqrt{T})$, where $d$ is the dimensionality of the arm features and $T$ is the number of rounds. The \LinUCB\ algorithm \citep{li2010contextual} attains this upper bound ($\tilde{O}(d\sqrt{T})$) \cite{lattimore2020bandit} and is also a popular choice for applications such as news recommendation \citep{li2010contextual} due to it's strong empirical performance. Moreover, it was also adapted to the hybrid setting in \cite{li2010contextual}. As a result we choose it as one of our main candidate algorithms to study in this paper. The popular \SupLinUCB\ \citep{chu2011contextual} algorithm improves the dependence on $d$ and guarantees $\tilde{O}(\sqrt{dT}\log^{3/2}{K})$ regret where $K$ is the number of actions. A recent variant of the \SupLinUCB\ algorithm \citep{li2019nearly} improves this further to $\tilde{O}(\sqrt{dT}\log{K})$. While the dependence on $d$ improves, it is traded off with a logarithmic dependence on $K$. It has also been noted \citep{Kim23} that despite strong regret guarantees, \SupLinUCB\ and its variants \citep{Li2017, li2019nearly} explore excessively and perform worse in practice due to computational inefficiency. 
Recently, \cite{Kim23} developed a new algorithm \HyRan\ which does not depend on $K$ in the leading terms and attains a regret guarantee of $\tilde{O}(\sqrt{dT})$ (improving significantly over \SupLinUCB\ both theoretically and empirically). While they prove this under an additional assumption on stochasticity of the arm features, they also provide a matching lower bound showing tightness of their guarantee. Their experiments validate their strong upper bound by demonstrating superior performance compared to many state of the art \lincb\ baselines for a large variety of problem instances. Due to their strong guarantee and empirical performance, we compare our algorithms with \HyRan\ in Section \ref{section:experiments}.
Another family of algorithms (based on randomized exploration) referred to as Thompson Sampling \citep{Thompson1933} based methods are an active area of research. Linear contextual version of this method was recently developed \citep{Agarwal2013, Abeille17} and best known regret guarantees are either $\tilde{O}(d\sqrt{T\log{K}})$ or $\tilde{O}(d^{3/2}\sqrt{T})$. Both guarantees are worse than that of \LinUCB. Even though Thompson Sampling does well practically, the \HyRan\ algorithm from \cite{Kim23} was shown to be superior to them for all problem instances considered in \cite{Kim23}. Since we already compare with \HyRan\ in our experiments in Section \ref{section:experiments}, we do not compare to Thompson Sampling based methods.

%% file: formulation.tex
\section{Problem Formulation}
\label{section:formulation}


\textbf{Notations:} $[N]$ denotes the set $\{1,\ldots,N\}$. $K, d_1, d_2 \in \N$ denote the number of arms, shared parameters and arm-specific parameters respectively. We denote by $\lVert \cdot \rVert_2$ the $\ell_2$ norm of a vector and by $\lVert \cdot \rVert$ the operator norm of a matrix. We denote all matrices with boldface and vectors in small case. For two symmetric matrices $\Ab$ and $\Bb$, we write $\Ab \succcurlyeq \Bb$ to indicate that $\Ab-\Bb$ is positive semi-definite. Further, we will denote by $\Ib_p$ the $p \times p$ identity matrix and $\zero$ as the zero matrix/vector (whose dimension will be clear from context when not specified).

\textbf{\lincb\ with hybrid rewards:} We assume there exist unknown parameter vectors $\theta^* \in \R^{d_1}$ and $\beta_i^* \in \R^{d_2}$ for $i\in [K]$. At each round $t\in \N$, the learner receives a set of $K$ feature vector tuples $\cX_t = \{(x_{i,t}, z_{i,t}): i\in [K]\}$. For each arm $i\in [K]$, the features $x_{i,t}\in \R^{d_1}$ correspond to the parameter vector $\theta^*$ and the features $z_{i,t}\in \R^{d_2}$ correspond to the arm-specific parameters $\beta_i^*$, i.e. if the learner selects arm $i_t\in [K]$ at round $t$, she receives a reward
\begin{equation*}
    r_t = \langle x_{i_t,t}, \theta^*\rangle + \langle z_{i_t,t}, \beta_{i_t}^*\rangle + \eta_t,
\end{equation*}
where $\eta_t$ is a conditionally $1$-subgaussian random noise. Specifically, let $\cF_t$ be the filtration $(\cX_1, i_1, r_1, \dots, r_{t-1})$, then, 
$\EE[\eta_t \mid \cF_t] = 0,\text{ and, } \EE[e^{\alpha \eta_t}\mid \cF_t] \leq e^{\alpha^2 / 2}
$,    
for all $\alpha \in \R$. Note that we do not assume the noise variable $\eta_t$ to depend on the arms selected by the learner. This later allows us to reduce the problem to the shared setting albeit in a larger dimension. We define the (cumulative) regret of the learner \texttt{Alg} at the end of round $T$ as,
\begin{align*}
    \Reg(T, \texttt{Alg}) = &\sum\limits_{t=1}^T\max_{j \in [K]} \left(\dotp{x_{j,t}}{\theta^*} + \dotp{z_{j,t}}{\beta_j^*}\right) - \sum\limits_{t=1}^T\left(\dotp{x_{i_t,t}}{\theta^*} + \dotp{z_{i_t,t}}{\beta_{i_t}^*}\right).
\end{align*}

\textbf{Assumptions:} The key assumption in this work is the diversity assumption, similar to the one made in~\citep{chatterji2020osom, ghosh2021problem, gentile2017context} for the shared setting. We extend this assumption (stated below) to the hybrid setting by adding one extra assumption that arises from the hybrid nature of the problem.

\begin{assumption}[Diversity]
\label{assumption:independent-subgaussian-features}
There exists a constant $\rho > 0$, such that features of the arm selected by the algorithm, i.e., $x_{i_t, t}$, $z_{i_t, t}$, for all rounds $t\in \NN$, satisfy,
\begin{enumerate}
    \item $\EE[x_{i_t,t} \mid \cF_{t-1}] = \zero, \EE[z_{i_t,t} \mid \cF_{t-1}] = \zero,$ and $ \EE\sbr{x_{i_t,t} z_{i_t,t}^\intercal \mid \cF_{t-1}} = \zero$,
    \item $\EE\sbr{x_{i_t,t} x_{i_t,t}^\intercal \mid \cF_{t-1}} \succcurlyeq \rho \Ib_{d_1}, \EE\sbr{z_{i_t,t} z_{i_t,t}^\intercal \mid \cF_{t-1}} \succcurlyeq \rho \Ib_{d_2}$
\end{enumerate}
\end{assumption}

We note that this assumption is algorithm dependent, however, there are problem instances in the shared setting for which any algorithm satisfies it (Section $3$, \citep{pmlr-v139-papini21a}). We empirically validate this assumption in the hybrid setting for the algorithms studied in this work, thereby demonstrating that the assumption and its crucial implications indeed hold in practice. The details of this empirical study is presented in Appendix~\ref{appendix:empirical-validation}. We also make the following standard assumption on the reward parameters.
 
\begin{assumption}
\label{assumption:boundedness}
There is a fixed constant $S\in \R$ (known to us) such that the reward parameters $\norm{\thta}{2} \leq S$ and $\norm{\bta{i}}{2} \leq S$ for all $i \in [K]$. Further, $\norm{x_{i,t}}{2} \leq 1$ and $\norm{z_{i,t}}{2} \leq 1$, for all arms $i \in [K]$ and rounds $t\in \NN$.
\end{assumption}

\textbf{Reduction to shared setting:} We will now formalize our discussion from Section \ref{subsection:additional-remarks} about reducing the hybrid setting to the shared setting. We first embed the arm features into a sparse feature vector in $\R^{d_1+d_2K}$ via transformation $\cP:[K] \times \R^{d_1} \times \R^{d_2} \rightarrow \R^{d_1 + d_2 K}$ which takes as input arm $i\in [K]$ and its features $x\in \R^{d_1}, z\in \R^{d_2}$ and maps it as follows:
\begin{align}
    \cP(i, x, z) = (x^\intercal, 0, \ldots,0, z^\intercal, 0,\ldots,0)^\intercal
\end{align}
that is, the coordinate entries of $x$ are copied to the first $d_1$ coordinates of $\cP(i, x, z)$ while the coordinate entries of $z$ are copied to the location starting from $d_1 + (i-1)d_2 + 1$ and ending at $d_1 + i d_2$. We also combine the true shared and arm-specific parameters into a global parameter vector $\phi^* = (\theta^{*\intercal}, \beta_{1}^{*\intercal}, \ldots, \beta_{K}^{*\intercal})^\intercal$. It's easy to see that this transformation converts the hybrid setting to the shared setting while preserving the mean reward of every arm, i.e.,
\begin{equation*}
    \dotp{\cP(i,x,z)}{\phi^*} = \dotp{x}{\thta} + \dotp{z}{\bta{i}} \quad \forall\ i \in [K].
\end{equation*}
Moreover, since the noise variable $\eta_t$ at round $t$ is assumed to be independent of the arm selected, the noisy rewards $r_t = \dotp{\cP(i,x,z)}{\phi^*} + \eta_t$ in this shared setting exactly captures the reward received by arm $i$ in the hybrid setting.

\textbf{More Notations.} 
For a matrix $\Mb \in \RR^{p \times p}$ and vector $a \in \RR^p$, for any $p>0$, we define $\norm{a}{\Mb} = \sqrt{a^\intercal \Mb a}$. For a square matrix $\Ab \in \RR^{p \times p}$, we will denote by $\lambda_{max}(\Ab)$ and $\lambda_{min}(\Ab)$ its maximum and minimum eigenvalues respectively. Finally, for $p, q \in \RR$, we will write $p \vee q = \max\cbr{p, q}$ and $p \wedge q = \min\cbr{p, q}$.

%% file: algorithms.tex
\section{Algorithms and Analysis}
\label{section:algorithms}
In this section, we present our main algorithms and their analyses. First, in Section \ref{subsection:linucb-dislinucb} we present the \LinUCB\ and \DisLinUCB\ algorithms for the hybrid setting in Algorithms \ref{algo:hylinUCB} and \ref{algo:dislinUCB} respectively. These algorithms are well known, however, for completeness we present them with the minor modifications we need to make to apply them to the hybrid setting. Following this, in Theorem \ref{theorem:regret-bound-linucb} and Corollary \ref{corollary:regret-bound-DisLinUCB}, we present the improved regret guarantees for \LinUCB\ and \DisLinUCB\ (under Assumptions \ref{assumption:independent-subgaussian-features} and \ref{assumption:boundedness}) respectively. Next, in Section \ref{subsection:hylinucb}, we introduce the \HyLinUCB\ algorithm which is exactly the same as \LinUCB\ presented in Algorithm \ref{algo:hylinUCB} but with a new confidence parameter $\gamma$ that we describe. We provide some intuition regarding our choice of this parameter and then in Theorem \ref{theorem:regret-bound-hylinucb} present a regret guarantee for this algorithm under Assumptions \ref{assumption:independent-subgaussian-features} and \ref{assumption:boundedness}. 

\subsection{\LinUCB\ and \DisLinUCB}
\label{subsection:linucb-dislinucb}

The \LinUCB\ algorithm (Algorithm \ref{algo:hylinUCB} with appropriate choices described below) takes as input the total number of rounds $T$, a regularization parameter $\lambda$, an exploration coefficient $\gamma$ (whose exact values we provide later) that controls the weightage to be put on the UCB (Upper Confidence Bound) bonus term, and a desired failure probability $\delta\in (0,1)$. In \emph{Step 1}, it performs initialization of the reward parameter vector and the design matrix $\Mb$. Note that all initialization is done assuming a feature space of dimension $d_1+d_2K$ which is the dimensionality after reducing the hybrid setting to the shared setting. The algorithm runs for rounds $t=1$ to $t=T$ (\emph{Steps 2-9}).
In \emph{Steps 3,4}, it receives the set $\cX_t$ of arm features and for each arm $i$, converts its features to appropriate shared setting features $\tilde{x}_i\in \R^{d_1+d_2K}$, using the mapping $\cP$ described in Section \ref{section:formulation}. Then, in \emph{Step 5}, it selects the arm $i_t$ with the best UCB estimate, and receives a reward $r_t$ corresponding to it. This is then used to update the design matrix and the estimated reward parameters in \emph{Steps 8,9}. For our \LinUCB\ implementation, we set the regularizer $\lambda=1$ and carefully choose the exploration coefficient $\gamma =  S\sqrt{K} + \sqrt{2(d_1 + d_2 K)\log(T/\delta)}$\footnote{$S$ is a known upper bound on the magnitude of the reward parameter vector} (in accordance with Lemma~\ref{lemma:theta-tilde-theta-bar-under-M-norm-confidence-set-guarantee} in \cite{abbasi2011improved}). This choice of $\lambda$ and $\gamma$ is standard for \LinUCB\ in the $d_1+d_2K$ dimensional shared setting. For these values of $\lambda$ and $ \gamma$, the regret analysis of \LinUCB(\citep{li2010contextual, abbasi2011improved, lattimore2020bandit}), under Assumption \ref{assumption:boundedness}, leads to a regret guarantee of $\tilde{O}((d_1+d_2K)\sqrt{T})$. In Theorem \ref{theorem:regret-bound-linucb}, we present a much stronger guarantee when Assumption \ref{assumption:independent-subgaussian-features} is also satisfied.

\begin{algorithm}[ht]
\caption{\LinUCB($\lambda, \gamma$)}
\label{algo:hylinUCB}
\begin{algorithmic}[1]
    \REQUIRE Regularizer $\lambda$, exploration coefficient $\gamma$, failure probability $\delta \in (0, 1)$
    \STATE Initialize $\thtill = \zero \in \RR^{d_1 + d_2 K}$, $\Mb = \lambda \Ib_{d_1 + d_2 K}$, $u = \zero \in \RR^{d_1 + d_2 K}$
    \FOR{$t=1,\dots,T$}
        \STATE Receive context $\cX_t = \{(x_i, z_i)\}_{i \in [K]}$
        \STATE $\Tilde{x}_i \coloneqq \cP(i,x_i,z_i)$ for all $i \in [K]$
        \STATE $i_t = \argmax_{i \in [K]} \dotp{\Tilde{x}_i}{\thtill} + \gamma \norm{\Tilde{x}_i}{\Mb^{-1}} $
        \STATE Play arm $i_t$ and observe reward $r_t$
        \STATE $u \gets u + r_t \Tilde{x}_t$
        \STATE $\Mb \gets \Mb + \Tilde{x}_t \Tilde{x}_t^\intercal$
        \STATE Update: $\thtill = \Mb^{-1} u$
        \alglinelabel{line:update-estimator}
    \ENDFOR
\end{algorithmic}
\end{algorithm}

\begin{theorem}[Regret of \LinUCB]
\label{theorem:regret-bound-linucb}
    At the end of $T$ rounds, the regret of \LinUCB\ (Algorithm \ref{algo:hylinUCB} with $\lambda=1$, $\gamma = S \sqrt{K} + \sqrt{2(d_1 + d_2 K)\log(T/\delta)}$) under Assumptions~\ref{assumption:independent-subgaussian-features} and \ref{assumption:boundedness} is upper bounded by $ C \sqrt{(d_1 + d_2 K) K T \log(T/\delta)}$ with probability at least $1 - 4\delta$. Here, $C > 0$ is a universal constant and $T$ is assumed to be $\tilde{\Omega}(K^4)$.
\end{theorem}
\noindent
The reduction from the hybrid setting to the shared setting leads to a sparse design matrix. Our proof of Theorem \ref{theorem:regret-bound-linucb} critically exploits the structure of this sparse matrix and its inverse, along with some key technical ideas involving geometry of random vectors and eigenvalue bounds on random matrices applied to the stochastic arm features and the design matrix. For brevity, we present the complete proof in Appendix \ref{appendix:main-proofs-from-sec-3}. 



Our second algorithm \DisLinUCB\ presented in Algorithm \ref{algo:dislinUCB} (same as Algorithm $1$ in \citep{li2010contextual}) takes the same set of inputs as Algorithm \ref{algo:hylinUCB} above, i.e., the total number of rounds $T$, a regularization parameter $\lambda$, an exploration coefficient $\gamma$ and a desired failure probability $\delta\in (0,1)$. In \emph{Step 1}, for each arm $i\in [K]$, it performs initialization of the reward parameter vectors $\phi_i$ and $(d_1+d_2)\times (d_1+d_2)$ size design matrices $\Vb_i$. This algorithm treats the hybrid setting as a disjoint setting and therefore the dimensionality of the problem remains the same as the hybrid setting i.e. $d_1+d_2$. The algorithm runs for rounds $t=1$ to $t=T$ (\emph{Steps 2-9}).
In \emph{Steps 3,4}, it receives the set $\cX_t$ of arm features. For each arm $i\in [K]$, it concatenates the shared and disjoint arm features to obtain the $d_1+d_2$ dimensional feature vector $\overline{x}_i$ and treats it as the vector of arm-specific parameters in the algorithm. Then, in \emph{Step 5}, it selects the arm $i_t$ with the best UCB estimate, and receives a reward $r_t$ corresponding to it. This is then used to update the corresponding design matrix (i.e. $\Vb_{i_t}$) and the estimated reward parameter vector (i.e. $\phi_{i_t}$) in \emph{Steps 8,9}. A key difference from the previous algorithm is that in \DisLinUCB\ the UCB bonus for arm $i\in [K]$ is computed using the arm specific design matrix i.e. $\Vb_i$, whereas in \LinUCB\, it is computed using the overall design matrix $\Mb$. In our \DisLinUCB\ instantiation, we set the regularizer $\lambda=1$ and choose the exploration coefficient $\gamma = 2\sqrt{S} + \sqrt{2(d_1 + d_2) \log(KT/\delta)}$. This choice of $\lambda$ and $\gamma$ is standard for \LinUCB\ in the $d_1+d_2$ dimensional shared setting. For these values of $\lambda$ and $\gamma$, the regret analysis of \LinUCB\ (under Assumption \ref{assumption:boundedness}) can be appropriately modified to provide a regret guarantee of $\tilde{O}((d_1+d_2)\sqrt{KT})$ for \DisLinUCB. In Corollary \ref{corollary:regret-bound-DisLinUCB}, we present a much stronger guarantee when Assumption \ref{assumption:independent-subgaussian-features} is also satisfied.

\begin{algorithm}[ht]
\caption{\DisLinUCB (Algorithm $1$ in \citep{li2010contextual})}
\label{algo:dislinUCB}
\begin{algorithmic}[1]
    \REQUIRE Regularizer $\lambda$, exploration coefficient $\gamma$, failure probability $\delta \in (0, 1)$
    \STATE For every arm $i \in [K]$, initialize $\phi_i = \zero \in \RR^{d_1 + d_2}$, $\Vb_i = \lambda \Ib_{d_1 + d_2}$, $u_i = \zero \in \RR^{d_1 + d_2}$
    \FOR{$t = 1, \dots, T$}
    \STATE Receive context $\cX_t = \{(x_i, z_i)\}_{i \in [K]}$
    \STATE Set $\overline{x}_i = \begin{bmatrix}
        x_i^\intercal & z_i^\intercal
    \end{bmatrix}^\intercal$ for all $i \in [K]$
    \STATE $i_t = \argmax_{i \in [K]} \dotp{\overline{x}_i}{\phi_i} + \gamma \norm{\overline{x}_i}{\Vb_i^{-1}}$
    \STATE Play arm $i_t$ and observe reward $r_t$
    \STATE $u_{i_t} \gets u_{i_t} + r_t \overline{x}_{i_t}$
    \STATE $\Vb_{i_t} \gets \Vb_{i_t} + \overline{x}_{i_t} \overline{x}_{i_t}^\intercal$
    \STATE Update $\phi_{i_t} = \Vb_{i_t}^{-1} u_{i_t}$
    \ENDFOR
\end{algorithmic}
\end{algorithm}

\begin{corollary}[Regret of \DisLinUCB]
\label{corollary:regret-bound-DisLinUCB}
At the end of $T$ rounds, the regret of \DisLinUCB\ (Algorithm \ref{algo:dislinUCB} with $\lambda=1$, $\gamma = 2\sqrt{S} + \sqrt{2(d_1 + d_2) \log(KT/\delta)}$) under Assumptions \ref{assumption:independent-subgaussian-features} and \ref{assumption:boundedness} is upper bounded by $C \sqrt{(d_1 + d_2) K T \log(K T / \delta)}$ with probability at least $1 - 4\delta$. Here, $C > 0$ is a universal constant and $T$ is assumed to be $\tilde{\Omega}(1)$. 
\end{corollary}
\noindent
Proof of Corollary \ref{corollary:regret-bound-DisLinUCB} uses techniques similar to the ones used in the proof of Theorem \ref{theorem:regret-bound-linucb} with modifications to adapt it to the disjoint setting. Complete proof is provided in Appendix \ref{appendix:dislinucb-regret-analysis}.

\subsection{\HyLinUCB}
\label{subsection:hylinucb}
Our new algorithm \HyLinUCB, is derived from Algorithm \ref{algo:hylinUCB} by using different values for parameters $\lambda$ and $\gamma$. In our \HyLinUCB\ instantiation, we set the regularizer $\lambda=K$ and choose the exploration coefficient $\gamma = 2(S\sqrt{K} + \sqrt{2(d_1 + d_2) \log(T/\delta))}$. The rest of the algorithm is exactly the same as Algorithm \ref{algo:hylinUCB}. However, due to this change the known regret analysis of \LinUCB\ (under Assumption \ref{assumption:boundedness}) cannot be directly applied to get optimal regret with respect to $T$. The optimality of the guarantee relies heavily on choosing $\gamma$ correctly (since it represents confidence in estimation of model parameters) and therefore, even simple modifications do not seem to work. Our main intuition behind using this new value for $\gamma$ is that even though the feature vector post reduction to the shared setting i.e. $\tilde{x}_i$ (See Step $4$ in Algorithm \ref{algo:hylinUCB}) is $d_1 + d_2 K$ dimensional, it has only $d_1 + d_2$ non-zero entries. This leads us to the question whether we can use a tighter exploration parameter that depends on the true intrinsic dimensionality of our feature vectors. In Theorem \ref{theorem:regret-bound-hylinucb}, we prove a strong regret guarantee (optimal with respect to $T$) when Assumptions \ref{assumption:independent-subgaussian-features} and \ref{assumption:boundedness} are satisfied. We also provide strong empirical evidence in support of our choice for $\gamma$ in Section \ref{section:experiments}.

\begin{theorem}[Regret of \HyLinUCB]
\label{theorem:regret-bound-hylinucb}
    At the end of $T$ rounds, the regret of \HyLinUCB\ (Algorithm~\ref{algo:hylinUCB} with $\lambda=K$, $\gamma = 2(S\sqrt{K} + \sqrt{2(d_1 + d_2)\log(T/\delta)})$) under Assumptions \ref{assumption:independent-subgaussian-features} and \ref{assumption:boundedness} is upper bounded by 
    \[
     C_1 \sqrt{K^3 T \log({K(d_1 + d_2)}/{\delta})} + C_2 \sqrt{(d_1 + d_2 K) K T \log({K T (d_1 + d_2)}/{\delta})}
    \]
with probability at least $1 - 4\delta$. Here $C_1, C_2 > 0$ are universal constants and $T$ is assumed to be $\tilde{\Omega}(K^4)$.
\end{theorem}
\noindent
Our proof of Theorem \ref{theorem:regret-bound-hylinucb} requires many new ideas on top of the techniques used in the proof of Theorem \ref{theorem:regret-bound-linucb} in order to accommodate the new confidence parameter. Complete proof is provided in Appendix \ref{appendix:hylinucb-regret-analysis}. We finish this section with a few remarks on the algorithm.

\begin{remark}
    The scheme of \HyLinUCB\ is similar to Algorithm $2$ in~\citep{li2010contextual}. However, in~\citep{li2010contextual}, the exploration coefficient $\gamma$ is treated as a hyperparameter to be tuned. As a result it's extremely difficult to compute a regret guarantee and no such guarantee is known. Moreover, it also brings a practical overhead of experimentally finding the optimal parameter. However, in \HyLinUCB, we fix an appropriate value of $\gamma$ and prove optimal (with respect to $T$) regret guarantees. This is where our algorithm differs from~\citep{li2010contextual}.
\end{remark}

\begin{remark}
    Line~\ref{line:update-estimator} in Algorithm~\ref{algo:hylinUCB} requires an inverse of a square matrix of $d_1 + d_2 K$ dimension, which requires $O((d_1 + d_2 K)^2)$ computation if one uses the Sherman-Morrison formula (Section $0.7.4$ in \cite{horn2012matrix}). This is still quadratic in $K$. However, it can be made linear in $K$ using block matrix inversion technique (see~\citep{li2010contextual,Stanimirovi2019InversionAP}). For simplicity, we present the algorithm in the above format.
\end{remark}

%% file: datasets.tex
\section{Experimental Setup}
\label{section:experiments}
In this section, we provide details of our experimental setup. We perform extensive experiments under different problem settings and on real world datasets to compare \HyLinUCB, \LinUCB, \DisLinUCB and a baseline \HyRan\ \citep{Kim23}. We set $\HyRan$ as our baseline because for the shared setting with similar assumptions, it was recently shown to have much better regret than all popular \lincb\ baselines described in Section \ref{section:related-work}. Our experiments are divided into two types (a) synthetic (b) real-world. We provide further details about the setup below. The code for all the experiments are available at \url{https://github.com/nirjhar-das/HyPay_Bandits}.

\subsection{Synthetic}
 For the synthetic experiments, the contextual arm features are generated by sampling randomly from some distribution. Since the hybrid reward setting is characterized by $3$ key parameters: $d_1$, $d_2$ and $K$, to be exhaustive in our evaluation, we design $3$ parameter settings to study the effect of these parameters on the regret of the various algorithms.
\paragraph{Parameter Settings: }
In \textbf{Setting 1}, we fix $d_1=40$ and $d_2=5$ to observe the regret when $d_2$ is small compared to $d_1$. In \textbf{Setting 2}, we reverse the situation and fix $d_1=5$ and $d_2=40$. For both \textbf{Setting 1} and \textbf{Setting 2} we fix $K=25$ and $T = 80,000$. For \textbf{Setting 3}, we fix $d_1=d_2=5$ and $T=30,000$ but vary $K$ over a set of points between $10$ and $400$. This helps us capture the dependence of regret on the number of arms $K$ for the various algorithms.

\paragraph{Environments:} We model the environment as a sequence of $T$ contexts (arm features) and a set of reward parameters. For each of our parameter settings, we create $5$ different environments. For each environment, we run $5$ parallel trials. In two different trials of the same environment, the sequence of contexts and the reward parameters remain unchanged, but the random noise in reward generation is allowed to vary.  

\paragraph{Stochastic Feature Generation: }
 For each environment, the contexts i.e., the shared and arm-specific features of an arm in a particular round are generated by uniformly sampling $d_1$ and $d_2$ sized vectors from unit balls in $d_1$ and $d_2$ dimensions respectively. 

\paragraph{Reward Simulation: }
Reward parameters for an environment are created by uniformly sampling the shared parameter $\theta^*$ from the $d_1$-dimensional ball of radius $1$ (i.e., $S=1$). Thereafter, $K$ vectors are sampled uniformly from the $d_2$-dimensional ball of radius $1$, which become the disjoint parameters $\{\beta_i^*\}_{i \in [K]}$.
The stochastic reward generated for arm $i \in [K]$ with features $(x_i,z_i)$ is $$r_t = \dotp{x_i}{\thta} + \dotp{z_i}{\bta{i}} + \eta~,$$ where noise $\eta$ is sampled from the Gaussian distribution $\cN(0,0.01)$.

\subsection{Real-World}
We also perform experiments on the Yahoo! Front Page Dataset~\citep{yahoo-dataset}. This dataset contains click logs from real time user interactions with the news articles displayed in the \emph{Featured tab} of the \emph{Today module} on Yahoo! Front Page during the first $10$ days of May $2009$. This dataset was also used in~\citep{li2010contextual} for experiments with the hybrid model.

\paragraph{Dataset Description: }
At every round, a user arrives and is presented with $K = 20$ news articles. The user is described by $6$ features and every article presented also has $6$ features. The first feature is always $1$ while the remaining $5$ features correspond to the $5$ membership features constructed via conjoint analysis with a bi-linear model as described in~\citep{chu2009ACS}. The dataset contains about $45$ million such user-article interaction entries. Further, one of the articles out of these $K$ articles is actually shown to the user, for which click (or no-click) is recorded.

\paragraph{Feature Construction: } At a particular round, suppose the user features are denoted by $u \in \RR^6$ and the arm (news article) features of arm $i\in [K]$ are denoted by $v_i \in \RR^6$. Then, for arm $i$, the shared features $x_i$ are given by vectorizing $u v_i^\intercal$, while the disjoint features $z_i$ are set equal to $v_i$. Thus, $d_1 = 36$ and $d_2 = 6$.

\paragraph{Parameter Learning: }
We first learn a reward model by training a hybrid linear regression model on the first $1M$ data points (contained in May 1, 2009 dataset). For every user and the article that was actually shown to the user, we create the shared and the arm parameters $x_i$ and $z_i$ (respectively) of the arm (article). Note that the index of the arm shown in round $n$, i.e., $i_n$, is also an essential part of this dataset (as we need to train the arm parameters too). The target variable $y_n$ is the click feedback, which is a boolean variable. Thus, for $N = 1M$ data points, we have $(x_{i_n,n}, z_{i_n, n}, y_n)_{n=1}^N$. Finally we obtain the parameters as:
\begin{align*}
    \thta, \{\bta{i}\}_{i \in [K]} = \argmin_{\theta, \{\beta_i\}_{i \in [K]}} \sum_{n=1}^N \rbr{\dotp{x_{i_n,n}}{\theta} + \dotp{z_{i_n, n}}{\beta_{i_n}} - y_n}^2
\end{align*}

\paragraph{Reward Simulation: }
For simulating the bandit experiments, the stochastic reward for an arm $i$ with features $x_i$, $z_i$ is generated as $$r_t = \dotp{x_i}{\thta} + \dotp{z_i}{\bta{i}} + \eta~,$$ where noise $\eta$ is sampled from the Gaussian distribution $\cN(0,0.0001)$. The reason for keeping the variance of the noise low is that it was observed that the mean rewards are themselves of the order $0.01$, so high noise variance causes the algorithms to take longer to demonstrate sub-linear regret due to smaller signal-to-noise ratio in each arm pull's feedback.

\section{Results}
\begin{figure}[tb]
    \centering
    \begin{subfigure}{0.49\textwidth}
        \includegraphics[ width=\textwidth]{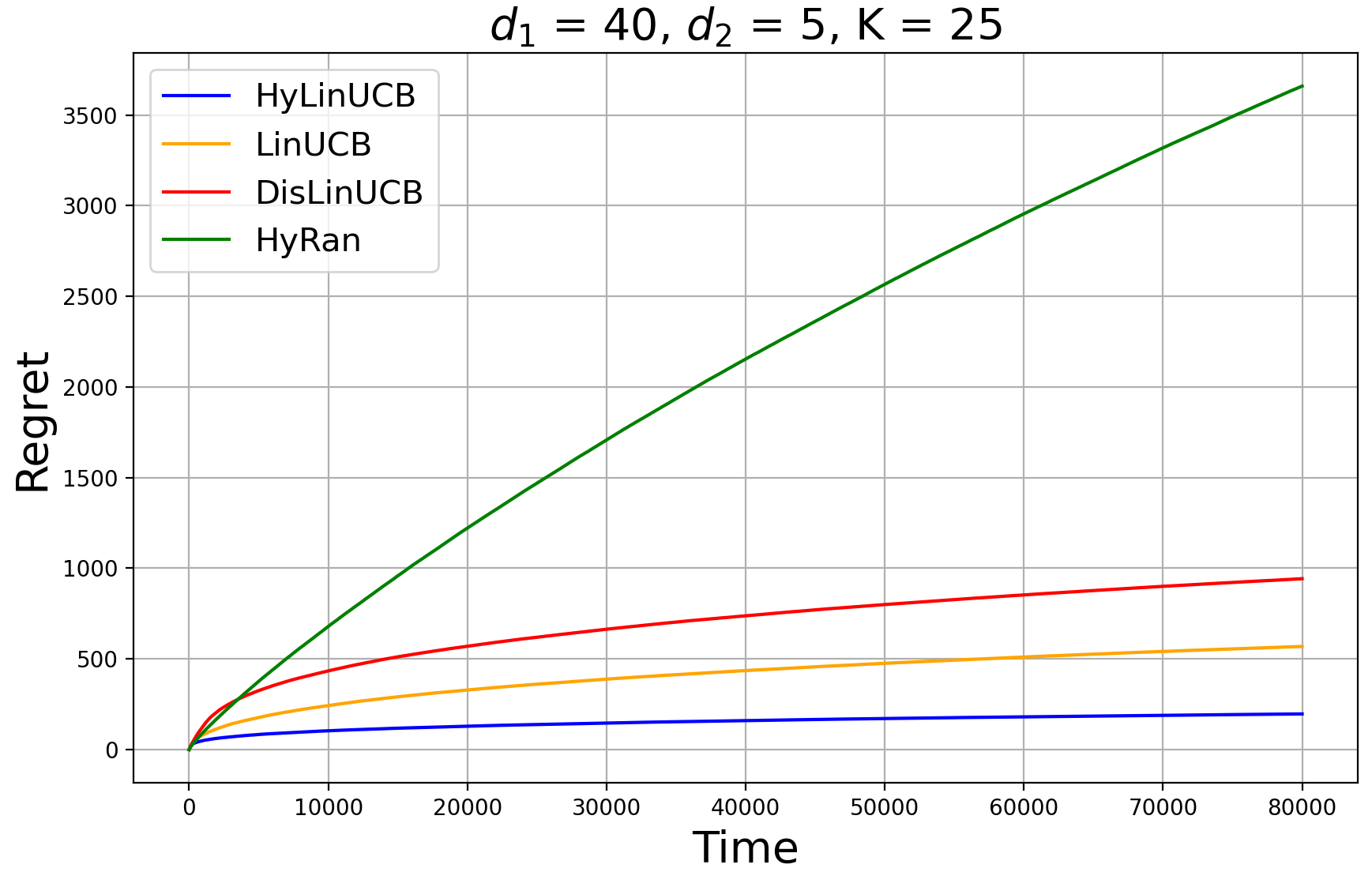}
        \label{subfig:reg-vs-T-d1=5-d2=40-K=25}
    \end{subfigure} 
    \hfill
    \begin{subfigure}{0.49\textwidth}
        \includegraphics[width=\textwidth]{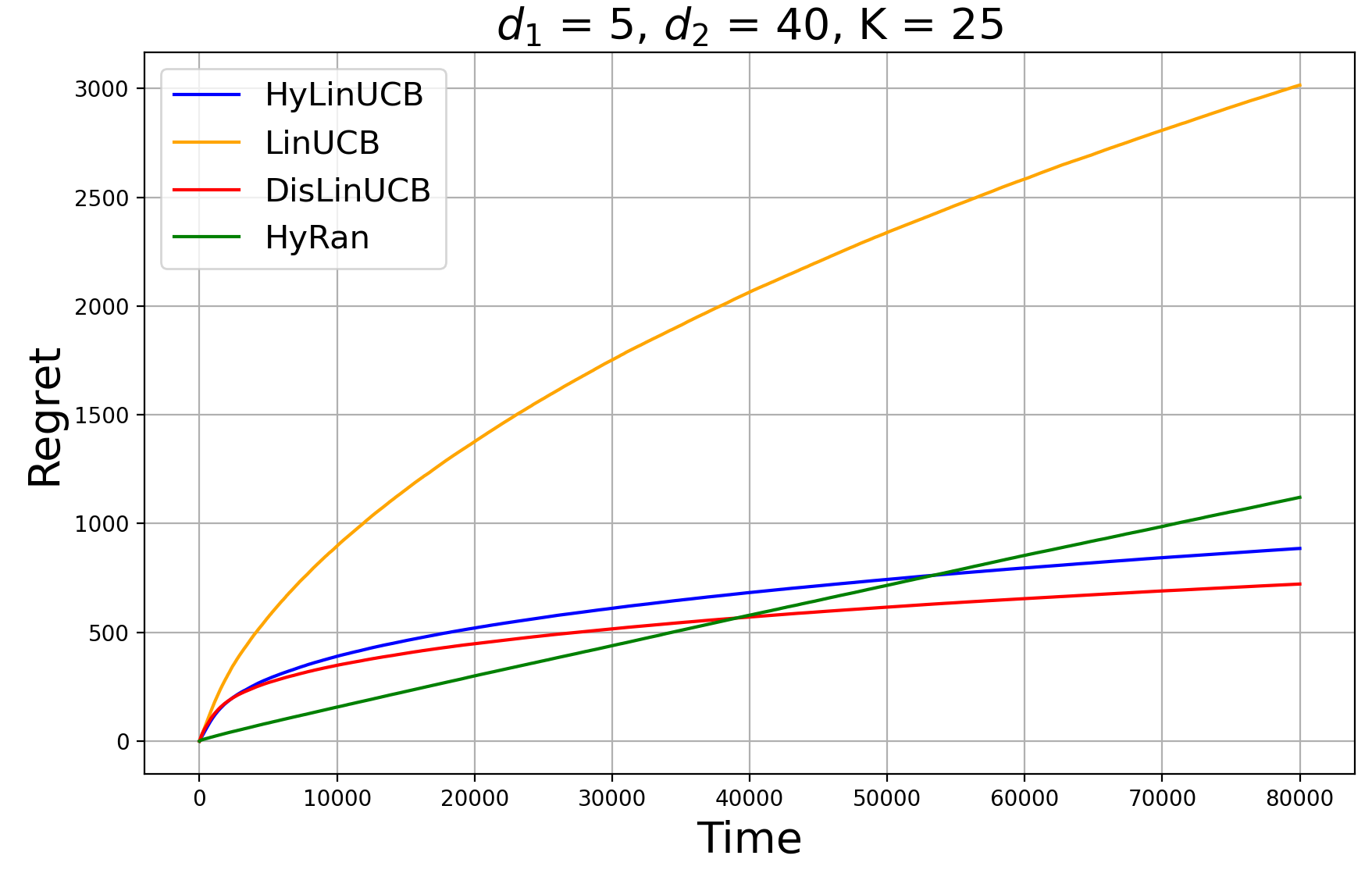}
        \label{subfig:reg-vs-T-d1=40-d2=5-K=25}
    \end{subfigure}
    \\
    \begin{subfigure}{0.49\textwidth}
        \includegraphics[width=\textwidth]{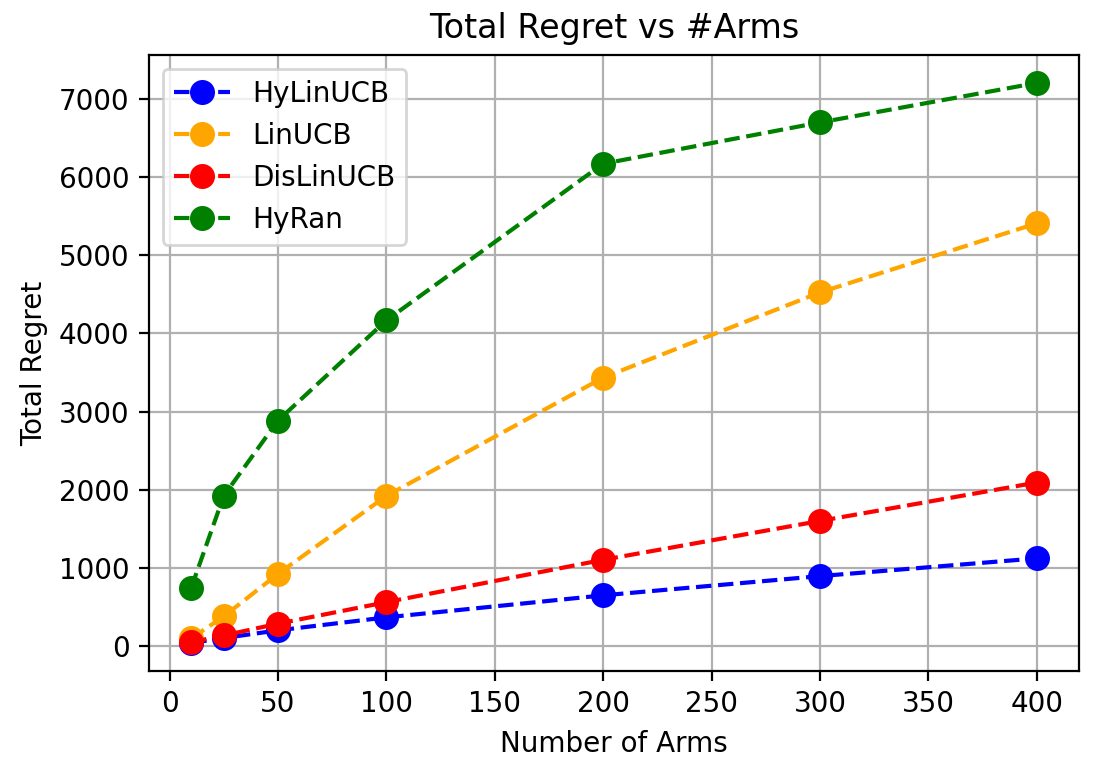}
        \label{subfig:reg-vs-num-arms}
    \end{subfigure} \hfill
    \begin{subfigure}{0.49\textwidth}
        \includegraphics[width=\textwidth]{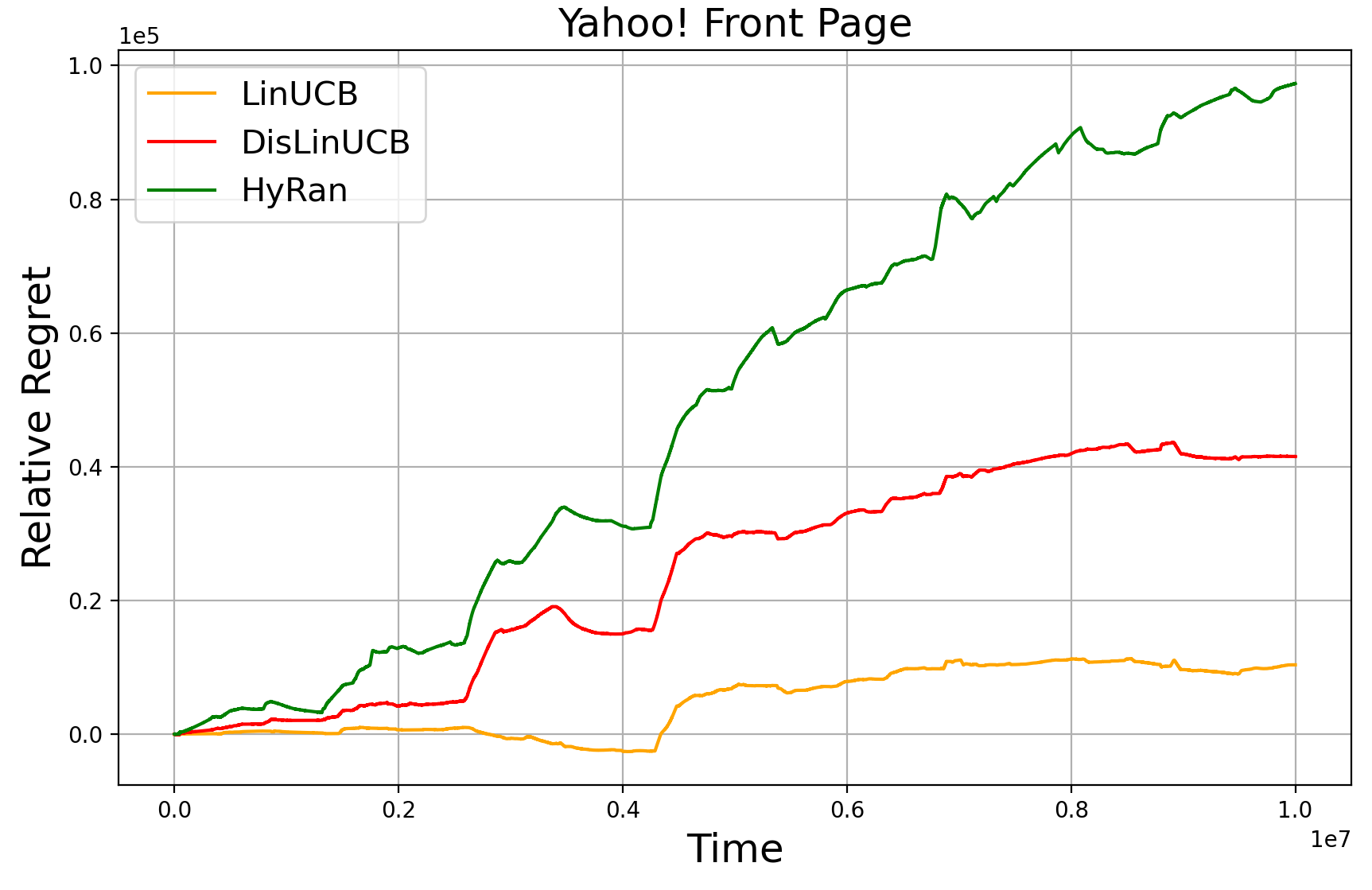}
        \label{subfig:yahoo-plot}
    \end{subfigure}
    \caption{Results of our experiments. Top-left: Regret vs \# of Rounds ($T$) for \textbf{Setting 1}; Top-right: Regret vs \# of Rounds ($T$) for \textbf{Setting 2}; Bottom-left: Regret versus \# of Arms for \textbf{Setting 3}; Bottom-right: Relative regret with respect to \HyLinUCB for Yahoo! Dataset.}
    \label{fig:result-fig}
\end{figure}
In all experiments, we compare \HyLinUCB, \LinUCB, \DisLinUCB and \texttt{HyRan}~\citep{Kim23}.
\subsection{Synthetic Experiments}

\paragraph{Regret vs $T$: }
In figure~\ref{fig:result-fig}, in the top row, we present results for \textbf{Setting 1} (left) and \textbf{Setting 2} (right). We plot the cumulative regret averaged over 5 parallel trials for each of the $5$ different environments. In \textbf{Setting 1}, with $d_1 \gg d_2$, we observe that the performance of \HyLinUCB is the best followed by \LinUCB, thus validating the superiority of these algorithms with higher shared parameters, which resembles more like a fully shared setting. \DisLinUCB also performs comparably but the regret is higher than \LinUCB. \texttt{HyRan} performs the worst in \textbf{Setting 1} although some sub-linear nature can be observed. In \textbf{Setting 2}, with $d_2 \gg d_1$, \DisLinUCB emerges as the best algorithm, while \HyLinUCB is a very close second. \HyRan exhibits a linear regret in this regime and eventually crosses beyond both \DisLinUCB and \HyLinUCB. Notably, \LinUCB has the worst performance in this setting, although the regret looks sub-linear. This matches with the intuition that \textbf{Setting 2} is closer to the fully disjoint setting, hence \LinUCB is not well-suited. However, the advantage of \HyLinUCB is very clear from these experiments as \HyLinUCB performs very well in both the settings, thus validating its suitability for the hybrid reward problem.

\paragraph{Regret vs $K$: }
In Fig.~\ref{fig:result-fig}, the left plot in the bottom row shows the effect of $K$ on the total regret. This plot corresponds to \textbf{Setting 3}, with $d_1 = d_2 = 5$. We again perform 5 parallel trials for (each of the) 5 different environments and then calculate the average of the total regret over $T$ rounds in all these $5\times 5 = 25$ simulations. The X-axis in the plot represents the number of arms $K$ while the Y-axis denotes the (average) total regret. From the plot, we can observe that \HyLinUCB has the smallest regret over all values of $K$ while \DisLinUCB is the second best, followed by \LinUCB and then \HyRan. \HyLinUCB displays a very slow growth with $K$, leading us to believe that its regret guarantee in Section \ref{section:algorithms} is not tight and can possibly be improved significantly.

\subsection{Real-World Experiment}
The results of the semi-synthetic experiment for $10M$ rounds (starting from May 2, 2009 dataset and moving to subsequent days' dataset till $10M$ rounds are complete) is shown in Fig.~\ref{fig:result-fig} in the bottom right. We plot the regret of other algorithms relative to \HyLinUCB (subtract the cumulative regret of \HyLinUCB from the cumulative regret of the algorithm), which has the smallest regret. \LinUCB comes close to \HyLinUCB\ whereas regret of \DisLinUCB\ and \HyRan\ are much larger. In this experiment, we do not have any control over the context features hence it is possible that the diversity assumption is not satisfied. However, we observe that \HyLinUCB still performs much better than the other algorithms, demonstrating strong evidence of good performance in hybrid models.

%% file: conclusion.tex
\section{Conclusion and Future Work}
\label{section:conclusion}
In this work, we revisited the problem of linear contextual bandits with hybrid rewards which is a useful setting in many applications such as news recommendation. The problem was originally proposed in~\citep{li2010contextual} albeit without any theoretical guarantees. Reducing this problem to the shared and disjoint settings, we derived regret guarantee for the \LinUCB and \DisLinUCB\ (Algorithm $1$ in \cite{li2010contextual}) algorithms, improving over their well known guarantees under a popular diversity assumption. Finally, we propose a new algorithm \HyLinUCB that employs a tighter exploration coefficient leveraging the sparsity of the problem. We also derive regret guarantee for this algorithm that has optimal dependence on $T$. We perform extensive empirical evaluation of the three algorithms in various synthetic scenarios (choices of $d_1, d_2$ and $K$) as well as on real-world datasets (Yahoo! Front Page Dataset \cite{yahoo-dataset}). We empirically compare these algorithms with the state-of-the-art \HyRan~\citep{Kim23} algorithm and demonstrate that \HyLinUCB outperforms the other algorithms in almost all cases and comes very close to the best in the remaining.

Since \HyLinUCB performs much better than \LinUCB empirically, an interesting future direction will be to derive tighter regret bounds that will provably demonstrate the efficacy of \HyLinUCB. Another future direction is to derive the regret guarantees under a different diversity assumption and understanding the trade-offs therein.

%% file: regret_linucb_appendix.tex
\section{Regret Analysis of \LinUCB}
\label{appendix:main-proofs-from-sec-3}

\paragraph{Notations} We first outline the notations of this section. For a rectangular matrix $\Db \in \RR^{p_1 \times p_2}$, we will denote by $\sigma_{max}(\Db)$ and $\sigma_{min}(\Db)$ the maximum and minimum singular values of $\Db$. Note that singular values of $\Db$ are given as the square root of the corresponding eigenvalue of $\Db^\intercal \Db$ (or $\Db \Db^\intercal$). Recall that in the \HyLinCB problem, the learner-environment interaction over $t$ rounds gives rise to the sequence of pulled arm feature tuples $\{(x_s, z_s)\}_{s=1}^t$ and the rewards $\{r_s\}_{s=1}^t$. Note that the problem can be converted to a \LinCB instance by thinking of the hybrid arm features actually coming from $\RR^{d_1  + d_2 K}$ as the modified vectors $\Tilde{x}_s = \cP(i_s, x_s, z_s)$ where $i_s$ is the index of the arm pulled in round $s$. We will denote by $\cT_{i,t}$ as the set of rounds till $t$ that the arm $i$ gets pulled and $\tau_{i,t} = \lvert \cT_{i,t} \rvert$, for all $i \in [K]$. Moreover, we will denote the failure probability with $\delta \in (0, 1)$ which is a fixed constant throughout. Let us first define three key matrices we will heavily use in this section.

\begin{align*}
    \Vb_t \coloneqq \lambda \Ib + \sum_{s=1}^t x_s x_s^\intercal && 
    \Bb_{i, t} \coloneqq \sum_{s \in \cT_{i, t}} x_s z_s^\intercal &&
    \Wb_{i, t} = \lambda \Ib + \sum_{s\in \cT_{i, t}} z_s z_s^\intercal
\end{align*}

Now recall the estimators for the hybrid model $\thth_t$ and $\{\bth{i, t}\}_{i=1}^K$ are given by:

\begin{align}
\label{eq:theta-hat-definition}
    \begin{bmatrix}
    \thth_t \\
    \bth{1, t} \\
    \vdots \\
    \bth{K, t}
    \end{bmatrix} = \Mb_t^{-1} \sum_{s=1}^t r_s \Tilde{x}_s \text{ where } \Mb_t \coloneqq \lambda \Ib + \sum_{s=1}^t \tilde{x}_s \tilde{x}_s^\intercal = \begin{bmatrix}
        \Vb_t & \Bb_{i, t} & \dots & \Bb_{K, t} \\
        \Bb^\intercal_{1, t} & \Wb_{1, t} & \dots & 0 \\
        \vdots & \vdots & \ddots & \vdots \\
        \Bb^\intercal_{K, t} & 0 & \dots & \Wb_{K, t}
    \end{bmatrix}
\end{align}
We will denote the vector of parameter estimates $\begin{bmatrix}
    {\thth_t}^\intercal & {\bth{1, t}}^\intercal &\dots & {\bth{K, t}}^\intercal
\end{bmatrix}^\intercal$ with the notation $\thtill_t$. Finally, we will define $\Ub_t$ which is the block diagonal part of $\Mb_t$ defined as follows:
\begin{align*}
    \Ub_t = \diag(\Vb_t, \Wb_{1, t}, \dots \Wb_{K, t})
\end{align*}

\subsection{Proof of Theorem~\ref{theorem:regret-bound-linucb}}
\label{appendix:proof-of-linucb}
After round $t$, from~\eqref{eq:theta-hat-definition} we have that $\Mb_t \thtill_t = \sum_{s=1}^t r_s \Tilde{x}_s$. Let $\thbar = \begin{bmatrix}
    {\thta}^\intercal & {\bta{1}}^\intercal & \dots & {\bta{K}}^\intercal
 \end{bmatrix}^\intercal$. By the definition of the reward model, we have for all $s \in [T]$, $r_s = \dotp{x_s}{\thta} + \dotp{z_s}{\bta{i_s}} + \eta_s$, where $\eta_s$ is the zero mean $1$-subgaussian noise and $i_s$ is the index of the arm pulled at round $s$.


\begin{lemma}
\label{lemma:theta-tilde-theta-bar-under-M-norm-confidence-set-guarantee}
    With probability at least $1 - \delta$, for all $t \geq 0$, $\norm{\thtill_t - \thbar}{\Mb_{t}} \leq \gamma \coloneqq 2\sqrt{\lambda K} S + \sqrt{2(d_1 + d_2 K)\log\left(T/\delta\right)}$.
\end{lemma}

\begin{proof}
    The proof is a direct application of Lemma~\ref{lemma:abbasi-yadkori-confidence-bound-of-estimator} with the fact that $\norm{\thbar}{2} = \sqrt{\norm{\thta}{2}^2 + \sum_{i=1}^K \norm{\bta{i}}{2}^2} \leq \sqrt{(K+1)S^2} = 2S \sqrt{K}$.
\end{proof}

\begin{lemma}
\label{lemma:estimation-error-for-any-vector-tilde-x}
    For any $\Tilde{x} \in \RR^{d_1 + d_2 K}$ and $t \geq 0$, with probability at least $1 - \delta$, we have $\lvert \dotp{\Tilde{x}}{\thtill_t - \thbar}\rvert \leq \gamma \norm{\Tilde{x}}{\Mb_t^{-1}} $.
\end{lemma}

\begin{proof}
    $\lvert \dotp{\Tilde{x}}{\thtill_t - \thbar}\rvert \leq \norm{\Tilde{x}}{\Mb_t^{-1}} \cdot \norm{\thtill_t - \thbar}{\Mb_t}$ by Cauchy-Schwarz inequality. Thereby, applying Lemma~\ref{lemma:theta-tilde-theta-bar-under-M-norm-confidence-set-guarantee} finishes the proof.
\end{proof}


We define $T_o$ and $T_m$ as: $T_m \coloneqq \rbr{\frac{16}{\rho^2} + \frac{8}{3\rho}} \log\rbr{\frac{2(d_1 + d_2) K T}{\delta}}$ and $$T_o \coloneqq \rbr{\frac{128}{\rho^2} \vee 4 T_m} K^2 \log\rbr{\frac{(d_1 + d_2) K}{\delta}}~.$$
We now define the following events which will be used to prove Theorem~\ref{theorem:regret-bound-linucb}. 
\begin{align*}
    \cE_1 &\coloneqq \cbr{\forall\ i \in [K], \forall\ t \geq 0: \norm{\Bb_{i,t}}{} \leq \sqrt{8 \tau_{i,t} \log\rbr{K(d_1 + d_2)/{\delta}}}} \\
    \cE_2 &\coloneqq \cbr{\forall\ t \geq T_m: \lambda_{min}\rbr{\Vb_t} \geq {\rho t}/{2}} \\
    \cE_3 &\coloneqq \cbr{\forall\ i \in [K], \forall\ t \geq 0 \text{ s.t. } \tau_{i,t} \geq T_m: \lambda_{min}\rbr{\Wb_{i,t}} \geq {\rho \tau_{i,t}}/{2}} \\
    \cE_4 &\coloneqq \cbr{\forall\ t \geq 0: \norm{\thtill_t - \thbar}{\Mb_{t}} \leq \gamma}
\end{align*}
Finally, we define $
    \cE \coloneqq \cE_1 \cap \cE_2 \cap \cE_3 \cap \cE_4$


\begin{lemma}
\label{lemma:condensed-lemma-for-linucb-proof}
    We have the following:
    \begin{enumerate}
        \item $\PP[\cE] \geq 1 - 4 \delta$
        \item Under event $\cE$, for all $t \geq T_o$, $\frac{1}{2} \Ub_t \preccurlyeq \Mb_t \preccurlyeq 2 \Ub_t$
    \end{enumerate}
\end{lemma}

\begin{proof}
    $\PP[\overline{\cE}] = \PP\sbr{(\overline{\cE_1 \cap \cE_2 \cap \cE_3}) \cup \overline{\cE_4}} \leq \PP\sbr{\overline{\cE_1 \cap \cE_2 \cap \cE_3}} + \PP\sbr{\overline{\cE_4}} \leq 3\delta + \delta = 4\delta$ where the first inequality is via Union bound and the second inequality is by application of Corollary~\ref{corollary:concentration-good-event-probability-linucb} and Lemma~\ref{lemma:theta-tilde-theta-bar-under-M-norm-confidence-set-guarantee}. Thus, $\PP[\cE] \geq 1 - 4 \delta$, which proves 1. Lastly, 2. is a restatement of Lemma~\ref{lemma:lowener-order-of-M-and-U} whose proof has been presented in Appendix~\ref{appendix:supporting-lemmas-for-LinUCB-analysis}.
\end{proof}

\begin{lemma}
\label{lemma:instantaneous-regret-bound}
    Let $i_t$ be the arm played in round $t > T_o$. Then, the instantaneous regret bound of \LinUCB is upper bounded by $\sqrt{8} \gamma \left(\norm{x_{t}}{\Vb_{t-1}^{-1}} + \norm{z_{t}}{\Wb_{i_t, t-1}^{-1}}\right)$ with probability at least $1 - 4\delta~.$
\end{lemma}

\begin{proof}
    Let $i^*_t$ denote the best arm in round $t$. Hereon, we will assume that the event $\cE$ holds, which happens with probability at least $1-4\delta~.$ For clarity, let $\thtill = \thtill_{t-1}$ The instantaneous regret in round $t$ is given by
    \begin{align*}
        \dotp{\Tilde{x}_{i^*_t,t}}{\thbar} - \dotp{\Tilde{x}_{i_t, t}}{\thbar}
        &\leq \dotp{\Tilde{x}_{i^*_t,t}}{\thtill} + \gamma \norm{\Tilde{x}_{i^*_t,t}}{\Mb_{t-1}^{-1}} - \dotp{x_{i_t,t}}{\thtill} + \gamma \norm{x_{i_t,t}}{\Mb_{t-1}^{-1}} \tag{by Lemma~\ref{lemma:estimation-error-for-any-vector-tilde-x}}\\
        &\leq \dotp{\Tilde{x}_{i_t,t}}{\thtill} + \gamma \norm{\Tilde{x}_{i_t,t}}{\Mb_{t-1}^{-1}}  - \dotp{x_{i_t,t}}{\thtill} + \gamma \norm{x_{i_t,t}}{\Mb_{t-1}^{-1}} \tag{$i_t = \argmax_{i \in [K]} \dotp{\Tilde{x}_{i,t}}{\thtill} + \gamma \norm{\Tilde{x}_{i,t}}{\Mb_{t-1}^{-1}}$}\\
        &= 2 \gamma \norm{\Tilde{x}_{i_t}}{\Mb_{t-1}^{-1}}\\
        &\leq 2 \sqrt{2} \gamma \norm{\Tilde{x}_{i_t,t}}{\Ub_{t-1}^{-1}} \tag{by Lemma~\ref{lemma:condensed-lemma-for-linucb-proof}}\\
        &= \sqrt{8} \gamma \sqrt{\norm{x_{i_t,t}}{\Vb_{t-1}^{-1}}^2 + \norm{z_{i_t}}{\Wb_{i_t, t-1}^{-1}}^2}\\
        &\leq \sqrt{8} \gamma \left( \norm{x_{i_t,t}}{\Vb_{t-1}^{-1}} + \norm{z_{i_t,t}}{\Wb_{i_t, t-1}^{-1}}\right) \tag{$\sqrt{a + b} \leq \sqrt{a} + \sqrt{b}$ for $a,b \geq 0$}
    \end{align*}
\end{proof}

\begin{restatable}{lemma}{ToRoundsLater}
\label{lemma:cumulative-regret-bound-after-T_o-rounds}
    Under event $\cE$, we have the following:
    \begin{align*}
        \sum_{t = T_o + 1}^T \left( \norm{x_{i_t,t}}{\Vb_{t-1}^{-1}} + \norm{z_{i_t,t}}{\Wb_{i_t, t-1}^{-1}}\right)
        &\leq 2 K \sqrt{d_2 T_m \log(T_m)} + \sqrt{\frac{32 K T}{\rho}}~.
    \end{align*}
\end{restatable}

\begin{proof}
    The proof of this Lemma is deferred till Appendix~\ref{appendix:supporting-lemmas-for-LinUCB-analysis}
\end{proof}

\begin{theorem}[Regret of \LinUCB]
\label{theorem:regret-bound-linucb_appendix}
    At the end of $T$ rounds, the regret of \LinUCB\ (Algorithm \ref{algo:hylinUCB} with $\lambda=1$, $\gamma = 2 S \sqrt{K} + \sqrt{2(d_1 + d_2 K)\log(T/\delta)}$) under Assumptions~\ref{assumption:independent-subgaussian-features}, \ref{assumption:boundedness} is upper bounded by $ C \sqrt{(d_1 + d_2 K) K T \log(T/\delta)}$ with probability at least $1 - 4\delta$. Here, $C > 0$ is a universal constant and $T$ is assumed to be $\tilde{\Omega}(K^4)$.
\end{theorem}

\begin{proof}
    The total regret after $T$ rounds is the sum of the instantaneous regrets over these rounds. Suppose $T > T_o$ (it is easy to see that a loose lower bound on $T_o$ is $\Tilde{\Omega}(K^4)$). Hereon, we will assume that the event $\cE$ holds, which happens with probability at least $1-4\delta~.$ Hence,
    \begin{align*}
        &\Reg(T, \LinUCB) \\
        &= \sum_{t=1}^T \dotp{\Tilde{x}_{i^*_t,t}}{\thbar} - \dotp{\Tilde{x}_{i_t,t}}{\thbar} \\
        &\leq 2 \gamma \sum_{t=1}^{T_o} \norm{\Tilde{x}_{i_t,t}}{\Mb^{-1}_{t-1}} + \sqrt{8} \gamma \sum_{t=T_o + 1}^T \left( \norm{x_{i_t,t}}{\Vb_{t-1}^{-1}} + \norm{z_{i_t,t}}{\Wb_{i_t, t-1}^{-1}}\right) \tag{by Lemma~\ref{lemma:instantaneous-regret-bound}}\\
        &\leq 2 \gamma \sqrt{4(d_1 + d_2 K)T_o \log(T_o)} +  \sqrt{8} \gamma \sum_{t=T_o + 1}^T \left( \norm{x_{i_t,t}}{\Vb_{t-1}^{-1}} + \norm{z_{i_t,t}}{\Wb_{i_t, t-1}^{-1}}\right) \tag{Lemma~\ref{lemma:elliptic-potential-lemma} on first term} \\
        &\leq 4 \gamma \sqrt{(d_1 + d_2 K)T_o \log(T_o)} + \sqrt{8} \gamma \rbr{2 K \sqrt{ d_2 T_m \log(T_m)} + \sqrt{\frac{32 K T}{\rho}}} \tag{Lemma~\ref{lemma:cumulative-regret-bound-after-T_o-rounds}}
    \end{align*}
    Finally, since we have $\gamma = 2 S \sqrt{\lambda K} + \sqrt{2 (d_1 + d_2 K) \log(T/\delta)}$, substituting in the above expression,
    \begin{align*}
    \Reg(T, \LinUCB) &\leq c_1 \rbr{2 S \sqrt{\lambda K} + \sqrt{2 (d_1 + d_2 K) \log(T/\delta)}} \\
    &+ 16\rbr{\frac{\sqrt{K}}{\sqrt{\rho}}} \rbr{2 S \sqrt{\lambda K} + \sqrt{2 (d_1 + d_2 K) \log(T/\delta)}} \sqrt{T}
    \end{align*}
    where $c_1 = 4\sqrt{(d_1 + d_2 K)T_o \log(T_o)} + 4 K \sqrt{2 d_2 T_m \log(T_m)}$. Simplifying constants and only keeping the asymptotically dominating terms in $T$,
    \begin{align*}
        \Reg(T, \LinUCB) \leq \frac{C}{\sqrt{\rho}} \sqrt{(d_1 + d_2 K) K T \log(T/\delta)}
    \end{align*}
    where $C > 0$ is a universal constant.
\end{proof}

\input{linucb_supp_lemmas_appendix}

%% file: linucb_supp_lemmas_appendix.tex
\subsection{Supporting Lemmas for~\ref{appendix:proof-of-linucb}}
\label{appendix:supporting-lemmas-for-LinUCB-analysis}

\begin{proposition}
\label{proposition:expression-for-M-(thtill-thbar)}
    $\Mb_t \left( \thtill_t - \thbar\right) = \sum_{s=1}^t \eta_s \Tilde{x}_s - \lambda \thbar$ for all $t \geq 0$.
\end{proposition}

\begin{proof}
    The mean reward of an arm $i \in [K]$ can also be written as $\dotp{\Tilde{x}_i}{\thbar}$. Note that $\Mb_t = \lambda \Ib + \sum_{s=1}^t \Tilde{x}_s \Tilde{x}_s^\intercal$. Consequently,
    \begin{align*}
        \Mb_t \thbar = \lambda \thbar + \sum_{S=1}^t \dotp{\Tilde{x}_s}{\thbar} \Tilde{x}_s
    \end{align*}
    Plugging this in the expression $\Mb_t \left(\thtill_t - \thbar \right)$ and using~\eqref{eq:theta-hat-definition}, we get:
    \begin{align*}
        \Mb_t \left( \thtill_t - \thbar \right) &= - \lambda \thbar + \sum_{s=1}^t r_s \Tilde{x}_s - \sum_{s=1}^t \dotp{\Tilde{x}_s}{\thbar} \Tilde{x}_s \\
        &= - \lambda \thbar + \sum_{s=1}^t \left(r_s - \dotp{\Tilde{x}_s}{\thbar}\right) \Tilde{x}_s
        = \sum_{s=1}^t \eta_s \Tilde{x}_s - \lambda \thbar
    \end{align*}
\end{proof}


\begin{proposition}
\label{proposition:relation-between-M-and-U-matrices}
    \begin{align*}
    \Ub^{-\frac{1}{2}}_t \Mb_t \Ub_t^{-\frac{1}{2}} &= \Ib + \begin{bmatrix}
        \zero & \Vb_t^{-\frac{1}{2}} \Bb_{1,t} \Wb_{1,t}^{-\frac{1}{2}} & \dots & \Vb_t^{-\frac{1}{2}} \Bb_{K,t} \Wb_{K,t}^{-\frac{1}{2}} \\
        \Wb_{1,t}^{-\frac{1}{2}} \Bb_{1,t}^\intercal \Vb_t^{-\frac{1}{2}}  & \zero & \dots & \zero\\
        \vdots & \vdots & \ddots & \vdots\\
        \Wb_{K,t}^{-\frac{1}{2}} \Bb_{K,t}^\intercal \Vb_t^{-\frac{1}{2}} & \zero & \dots & \zero
    \end{bmatrix}\\
    &\equiv \Ib + \Ab_t \tag{definition of $\Ab_t$}
\end{align*}
\end{proposition}

In what follows, we show that the maximum eigenvalue of $\Ab_t$ is bounded away from $1$ under the assumption~\ref{assumption:independent-subgaussian-features}, which in turn would imply Lemma~\ref{lemma:lowener-order-of-M-and-U}.

\begin{lemma}
\label{lemma:eigenvalue-properties-of-A-in-terms-of-Z}
    Let $\Zb_t = \begin{bmatrix}
        \Vb_t^{-\frac{1}{2}} \Bb_{1, t} \Wb_{1, t}^{-\frac{1}{2}} & \dots & \Vb_t^{-\frac{1}{2}} \Bb_{K,t} \Wb_{K, t}^{-\frac{1}{2}}
    \end{bmatrix}$. Then, it holds that almost surely, $\lambda_{max}(\Ab_t) = -\lambda_{min}(\Ab_t) = \sigma_{max}(\Zb_t)$.
\end{lemma}

\begin{proof}
    First, note that $\Ab_t =  \begin{bmatrix}
        \zero & \Zb_t \\
        \Zb_t^\intercal & \zero
    \end{bmatrix}$. Thus, $\Ab_t$ is the Hermitian dilation (see def.~\ref{definition-of-Hermitian-dilation}) of $\Zb_t$. Thus, by Lemma~\ref{lemma:hermitian-dilation-eigenvalue-relation}, the square of eigenvalues of $\Ab_t$ are equal to the square of the singular values of $\Zb_t$. Consequently, for every positive singular value $\sigma$ of $\Zb_t$, there are two eigenvalues of $\Ab_t$ namely, $\sigma$ and $-\sigma$. Hence, clearly, $\lambda_{max}(\Ab_t) = \sigma_{max}(\Zb_t)$ and $\lambda_{min}(\Ab_t) = - \sigma_{max}(\Zb_t)$.
\end{proof}

Now we turn to upper bounding $\norm{\Zb_t}{}$. Our goal is to show that $\norm{\Zb_t}{}$ is bounded away from $1$. First, we describe two high probability events for $t$ sufficiently large.

\begin{lemma}
\label{lemma:bound-on-max-norm-of-B}
    Let $\delta \in (0,1)$. Fix an $i \in [K]$. Then, for all $t \geq 0$, $\norm{\Bb_{i,t}}{} \leq \sqrt{8 \tau_{i,t} \log( (d_1 + d_2) K/\delta)}$ with probability at least $1 - \frac{\delta}{K}$.
\end{lemma}

\begin{proof}
    We have $\Bb_{i,t} = \sum_{s \in \cT_i} x_s z_s^\intercal$.
    By assumption~\ref{assumption:independent-subgaussian-features}, we have $\EE_{s-1} \sbr{x_s z_s^\intercal} = \zero~.$ Hence, $\Bb_{i,t}$ is a sum of Martingale difference sequences. Moreover, since $\norm{x_s}{2} \leq 1$ and $\norm{z_s}{2} \leq 1$, for the Hermitian dilation (definition~\ref{definition-of-Hermitian-dilation}) $\Hb_s$ of $x_s z_s^\intercal$,
    \begin{align*}
        \Hb_s^2 = \begin{bmatrix}
            \norm{z_s}{2}^2 x_s x_s^\intercal & \zero \\
            \zero & \norm{x_s}{2}^2 z_s z_s^\intercal
        \end{bmatrix} \preccurlyeq \Ib_{d_1 + d_2}
    \end{align*}
    Thus, by Matrix Azuma inequality (Lemma~\ref{lemma:matrix-azuma}), we can upper bound the operator norm of $\Bb_{i,t}$ as follows:
    \begin{align*}
        \PP\sbr{\exists t \geq 0: \norm{\Bb_{i,t}}{} \geq \varepsilon} \leq (d_1 + d_2)\exp\rbr{- \frac{\varepsilon^2}{8\tau_{i,t}^2}}
    \end{align*}
    Setting $\varepsilon = \sqrt{8 \tau_{i,t} \log(K(d_1 + d_2)/\delta)}$, we obtain,
    \begin{align*}
        \PP\sbr{\exists t \geq 0: \norm{\Bb_{i,t}}{} \geq \sqrt{8 \tau_{i,t} \log(K(d_1 + d_2)/\delta)}} \leq \frac{\delta}{K}
    \end{align*}
\end{proof}

\begin{corollary}
\label{corollary:concentration-of-B-matrix-for-all-i-and-all-t}
    For all $i \in [K]$ and $t \geq 0$, $\norm{\Bb_{i,t}}{} \leq \sqrt{8 \tau_{i,t} \log(K(d_1 + d_2)/\delta)}$ with probability at least $1 -\delta~.$
\end{corollary}
\begin{proof}
    Proof follows by applying union bound on Lemma~\ref{lemma:bound-on-max-norm-of-B}.
\end{proof}
Hereon, let us define the event $$\cE_1 \coloneqq \left\{\forall\ i \in [K], t \geq 0,\  \norm{\Bb_{i,t}}{} \leq \sqrt{8 \tau_{i,t} \log\rbr{\frac{K (d_1 + d_2)}{\delta}}}\right\}$$ Thus, we have that $\PP[\cE_1] \geq 1 - \delta$.

Next, we describe another event that happens with high probability under assumption~\ref{assumption:independent-subgaussian-features}.

\begin{lemma}
\label{lemma:min-eigenvalue-under-stochasticity-problem-based-result}
    We first define the quantity $T_m \coloneqq \rbr{\frac{16}{\rho^2} + \frac{8}{3\rho}} \log\rbr{\frac{2(d_1 + d_2) K T}{\delta}}$. Then, $\PP\sbr{\forall t \geq T_m: \lambda_{min}(\Vb_t) \geq \frac{\rho}{2}t} \geq 1 -\delta$, and $\PP\big[\forall i \in [K], \tau_{i,t} \geq T_m: \lambda_{min}(\Wb_{i,t}) \geq \frac{\rho}{2}\tau_{i,t} \big] \geq 1 -\delta$.
\end{lemma}

\begin{proof}
    The result is obtained by directly applying Lemma~\ref{lemma:min-eigenvalue-bartlett} followed by union bound.
\end{proof}

\noindent
Hereon, we will denote by $\cE_2$ as the event: $\{ \forall\ i\in [K],\ \forall\  \tau_{i,t} \geq T_m, \lambda_{min}(\Wb_{i,t}) \geq \rho \tau_{i,t}/2\}$ and by $\cE_3$ the event: $\{\forall\ t \geq T_m, \lambda_{min}(\Vb_t) \geq \rho t/2\}$. Finally, we will denote $$\cE_o \coloneqq \cE_1 \cap \cE_2 \cap \cE_3$$

\begin{corollary}
\label{corollary:concentration-good-event-probability-linucb}
    $\PP[\cE_o] \geq 1 - 3\delta~.$
\end{corollary}

\begin{proof}
    By Union Bound, $\PP[\overline{\cE_o}] \leq \PP[\overline{\cE_1}] + \PP[\overline{\cE_2}] + \PP[\overline{\cE_3}] \leq 3 \delta~.$
\end{proof}

\begin{lemma}
\label{lemma:norm-bound-of-Z}
    Under event $\cE_o$, for all $t > T_o \coloneqq \rbr{\frac{128}{\rho^2} \vee 4 T_m} K^2 \log\rbr{\frac{(d_1 + d_2) K}{\delta}}$, we have $\norm{\Zb_t}{} \leq \frac{1}{2}$
\end{lemma}

\begin{proof}
    Firstly, note that by definition, $\norm{\Zb_t}{} = \sup_{\norm{b}{2}=1} \norm{\Zb_t b}{2}$. Hence,
    \begin{align}
        \norm{\Zb_t}{} &= \sup_{\norm{b}{2}\leq 1} \norm{\Zb_t b}{2} \nonumber \\
        &= \sup_{\sum_{i=1}^K \norm{b_i}{2}^2 \leq 1} \norm{\sum_{i=1}^K \Vb_t^{-1/2} \Bb_{i,t} \Wb_{i,t}^{-1/2} b_i}{2} \tag{$b = [b_1\ b_2\ \dots\ b_K]^\intercal$} \nonumber\\
        &\leq  \sup_{\sum_{i=1}^K \norm{b_i}{2}^2 \leq 1} \sum_{i=1}^K \norm{\Vb_t^{-1/2} \Bb_{i,t} \Wb_{i,t}^{-1/2} b_i}{2} \tag{$\triangle$ inequality} \nonumber\\
        &\leq \sum_{i=1}^K \sup_{\norm{b_i}{2} \leq 1} \norm{\Vb_t^{-1/2} \Bb_{i,t} \Wb_{i,t}^{-1/2} b_i}{2} \tag{$\sup \sum \leq \sum \sup$; $b_i$ has larger domain} \nonumber \\
        &= \sum_{i=1}^K \norm{\Vb_t^{-1/2} \Bb_{i,t} \Wb_{i,t}^{-1/2}}{} \nonumber \\
        &\leq \sum_{i=1}^K \norm{\Vb_t^{-1/2}}{} \norm{\Bb_{i,t}}{} \norm{\Wb_{i,t}^{-1/2}}{} \tag{Sub-multiplicativity of norm} \nonumber \\
        &= \sum_{i=1}^K \frac{\norm{\Bb_{i,t}}{}}{\sqrt{\lambda_{min}(\Vb_t)} \sqrt{\lambda_{min}(\Wb_{i,t})}} \nonumber \tag{$\lambda_{\max}\left(\Db^{-1}\right) = \frac{1}{\lambda_{min}(\Db)}$}\\
        &\leq \sum_{i=1}^K \frac{\sqrt{8 \tau_{i,t} \log(K(d_1 + d_2)/\delta)}}{\sqrt{\lambda_{min}(\Vb_t)} \sqrt{\lambda_{min}(\Wb_{i,t})}}\label{eq:norm-Z-basic-sum}
    \end{align}
    We can split the RHS in~\eqref{eq:norm-Z-basic-sum} into two parts, one in which the arms have been pulled for less than $T_m$ times and the another containing the rest of the arms.
    Thus, we have 
    \begin{align}
    \norm{\Zb_t}{} \leq \sum_{i: \tau_{i,t} < T_m} \frac{\sqrt{8 \tau_{i,t} \log(K(d_1 + d_2)/\delta)}}{\sqrt{\lambda_{min}(\Vb_t)} \sqrt{\lambda_{min}(\Wb_{i,t})}} + \sum_{i: \tau_{i,t} \geq T_m} \frac{\sqrt{8 \tau_{i,t} \log(K(d_1 + d_2)/\delta)}}{\sqrt{\lambda_{min}(\Vb_t)} \sqrt{\lambda_{min}(\Wb_{i,t})}} \label{eq:norm-Z-into-two-parts}
    \end{align}
    
    We first bound the second term in~\eqref{eq:norm-Z-into-two-parts} as follows
    \begin{align*}
        \sum_{i: \tau_{i,t} \geq T_m} \frac{\sqrt{8 \tau_{i,t} \log(K(d_1 + d_2)/\delta)}}{\sqrt{\lambda_{min}(\Vb_t)} \sqrt{\lambda_{min}(\Wb_{i,t})}}
        &\leq \sum_{i: \tau_{i,t} \geq T_m} \frac{\sqrt{8 \tau_{i,t} \log (K (d_1 + d_2)/\delta)}}{\sqrt{(\rho t/2)} \sqrt{(\rho \tau_{i,t}/2)}} \tag{$\cE$ holds and $T_m \leq \tau_{i,t}$}\\
        &= \sum_{i: \tau_{i,t} \geq T_m} \frac{4 \sqrt{2\log(K(d_1 + d_2)/\delta)}}{\sqrt{\rho^2 t}} \\
        &\leq \sum_{i: \tau_{i,t} \geq T_m} \frac{1}{2 K} \tag{$t \geq \frac{128}{\rho^2} K^2 \log\rbr{\frac{(d_1 + d_2) K}{\delta}}$}
    \end{align*}
    Next, the first term in~\eqref{eq:norm-Z-into-two-parts} can be bounded as:
    \begin{align*}
        \sum_{i: \tau_{i,t} < T_m} \frac{\sqrt{8 \tau_{i,t} \log(K(d_1 + d_2)/\delta)}}{\sqrt{\lambda_{min}(\Vb_t)} \sqrt{\lambda_{min}(\Wb_{i,t})}} 
        &\leq \sum_{i: \tau_{i,t} < T_m} \frac{\sqrt{8 \tau_{i,t} \log(K(d_1 + d_2)/\delta)}}{\sqrt{\rho t / 2} \sqrt{\lambda}} \tag{$\cE$ holds; $\lambda_{min}(\Wb_{i,t}) \geq \lambda$}\\
        &\leq \sum_{i: \tau_{i,t} < T_m} \frac{4 \sqrt{T_m \log(K(d_1 + d_2)/\delta)}}{\sqrt{\rho \lambda t}}\\
        &\leq \sum_{i: \tau_{i,t} < T_m} \frac{1}{\sqrt{\lambda \rho}} \cdot \frac{1}{2 K} \tag{$t \geq 4 K^2 T_m \log(K(d_1 + d_2)/\delta)$}
    \end{align*}
    Choosing $\lambda \geq \frac{1}{\rho}$, we can have the RHS above smaller than $\frac{1}{2 K}$.
    Combining these results, we finally have,
    \begin{align*}
        \norm{\Zb_t}{} \leq  \sum_{i: \tau_{i,t} < T_m} \frac{1}{2 K} + \sum_{i: \tau_{i,t} \geq T_m} \frac{1}{2 K} = \frac{1}{2}
    \end{align*}
\end{proof}

\begin{lemma}
\label{lemma:lowener-order-of-M-and-U}
    Under event $\cE_o$, for all $t \geq T_o$, $\frac{1}{2}\Ub_t \preccurlyeq \Mb_t \preccurlyeq 2\Ub_t$.
\end{lemma}

\begin{proof}
    We first note that by Lemma~\ref{lemma:eigenvalue-properties-of-A-in-terms-of-Z}, $\lambda_{max}(\Ab_t) = - \lambda_{min}(\Ab_t)$. Thus, under $\cE_o$, $\lambda_{max}(\Ab_t) = \norm{\Zb_t}{} \leq \frac{1}{2}$ implies that $\lambda_{min}(\Ab_t) \geq -\frac{1}{2}$. Thus, all eigenvalues of $\Ab_t$ lie in the range $[-1/2, 1/2]$. This in turn implies that the eigenvalues of $\Ib + \Ab_t$ lie in the range $[1/2, 3/2]$. From this, it can be concluded that, $$\frac{1}{2}\Ib \preccurlyeq \Ib + \Ab_t \preccurlyeq 2\Ib$$ as eigenvalues of $2\Ib$ dominate that of $\Ib + \Ab_t$ which in turn dominate that of $(1/2)\Ib$. Thus, we have, by combining with Proposition~\ref{proposition:relation-between-M-and-U-matrices}, $\frac{1}{2}\Ib \preccurlyeq \Ub_t^{-\frac{1}{2}} \Mb_t \Ub_t^{-\frac{1}{2}} \preccurlyeq 2\Ib$, which then implies that,
    \begin{align*}
        \frac{1}{2}\Ub_t \preccurlyeq \Mb_t \preccurlyeq 2\Ub_t
    \end{align*}
\end{proof}

\ToRoundsLater*

\begin{proof}
    For a given $i \in [K]$, let the elements in $\cT_{i,T}$ be rearranged so that they are in increasing order. Let $\cT_{i,\xi}$ denote the subset of $\cT_{i,T}$ that contains only the first $T_m$ elements if $\abr{\cT_{i,T}} > T_m$ else $\cT_{i,\xi} = \cT_{i,T}$. Finally, let $\tau_{i,\xi} = \abr{\cT_{i,\xi}}$. Hence, the second term above can now be written as
    \begin{align*}
        \sum_{t=T_o}^T \norm{z_{i_t,t}}{\Wb_{i_t, t-1}^{-1}} &\leq \sum_{i=1}^K \sum_{s \in \cT_{i,T}} \norm{z_{i, s}}{\Wb_{i,s-1}^{-1}} \\
        &= \sum_{i=1}^K \rbr{\sum_{s \in \cT_{i,\xi}} \norm{z_{i, s}}{\Wb_{i,s-1}^{-1}} + \sum_{s \in \cT_{i,T} \setminus \cT_{i, \xi}} \norm{z_{i, s}}{\Wb_{i,s-1}^{-1}}}\\
        &\leq \sum_{i = 1}^K \sqrt{4 d_2 \tau_{i,\xi} \log(\tau_{i,\xi})} + \sum_{i=1}^K \sum_{s \in \cT_{i,T} \setminus \cT_{i, \xi}} \norm{z_{i, s}}{\Wb_{i,s-1}^{-1}} \tag{by Lemma~\ref{lemma:elliptic-potential-lemma}} \\
        &\leq \sum_{i = 1}^K \sqrt{4 d_2 T_m \log(T_m)} + \sum_{i=1}^K \sum_{s \in \cT_{i,T} \setminus \cT_{i, \xi}} \norm{z_{i, s}}{\Wb_{i,s-1}^{-1}}\tag{$\tau_{i,\xi} \leq T_m$} \\
        &\leq 2 K \sqrt{d_2 T_m \log(T_m)} + \sum_{i=1}^K \sum_{s \in \cT_{i,T} \setminus \cT_{i, \xi}} \frac{\norm{z_{i,s}}{2}}{\sqrt{\lambda_{min}(\Wb_{i,s-1})}} \\
        &\leq 2 K \sqrt{d_2 T_m \log(T_m)} + \sum_{i=1}^K \sum_{s \in \cT_{i,T} \setminus \cT_{i, \xi}} \frac{1}{\sqrt{\rho (\tau_{i,s}-1) / 2}} \tag{under $\cE$, $\lambda_{min}(\Wb_{i,t}) \geq \frac{\rho \tau_{i,t}}{2}$ since $\tau_{i,t} \geq T_m$; $\norm{z_{i,s}}{2} \leq 1$} \\
        &= 2 K \sqrt{d_2 T_m \log(T_m)} + \sum_{i=1}^K \ind{\tau_{i,T} > T_m} \sum_{s = T_m}^{\tau_{i,T}} \frac{1}{\sqrt{\rho (s-1) / 2}} \\
        &\leq 2 K \sqrt{d_2 T_m \log(T_m)} + \sum_{i=1}^K \ind{\tau_{i,T} > T_m} \sqrt{\frac{2}{\rho}} \int_{s = 1}^{\tau_{i,T}+1} \frac{1}{\sqrt{s}} \\
        &= 2 K \sqrt{d_2 T_m \log(T_m)} + \sum_{i=1}^K \ind{\tau_{i,T} > T_m} \sqrt{\frac{8 \tau_{i,T}}{\rho}} \\
        &\leq 2 K \sqrt{d_2 T_m \log(T_m)} + \sum_{i=1}^K \sqrt{\frac{8 \tau_{i,T}}{\rho}} \\
        &\leq 2 K \sqrt{d_2 T_m \log(T_m)} + \sqrt{\frac{8}{\rho}} \sqrt{K \cdot \sum_{i=1}^K \tau_{i,T}}
        \tag{Cauchy-Schwarz} \\
        &\leq 2 K \sqrt{d_2 T_m \log(T_m)} + \sqrt{\frac{8 K T}{\rho}} \tag{$\sum_{i=1}^K \tau_{i,T} = T$}
    \end{align*}
    Fnally, we have,
    \begin{align*}
        \sum_{t=T_o}^T \norm{x_{i_t}}{\Vb_{t-1}^{-1}} & \leq \sum_{t=T_o}^T \frac{\norm{x_{i_t}}{2}}{\sqrt{\lambda_{min}(\Vb_{t-1})}} \leq \sum_{t=T_o}^T \frac{1}{\sqrt{\rho (t-1) / 2}} \tag{by Lemma~\ref{lemma:condensed-lemma-for-linucb-proof}, $\lambda_{min}(\Vb_t) \geq \rho t/ 2$; $\norm{x_{i_t,t}}{2} \leq 1$} \\
        &\leq \sqrt{\frac{2}{\rho}} \int_{t=1}^{T+1} \frac{1}{\sqrt{t}} \leq \sqrt{\frac{8 T}{\rho}}
    \end{align*}
    Putting things back,
    \begin{align*}
        \sum_{t = T_o}^T \left( \norm{x_{i_t,t}}{\Vb_{t-1}^{-1}} + \norm{z_{i_t,t}}{\Wb_{i_t, t-1}^{-1}}\right) &\leq \sqrt{\frac{8 T}{\rho}} + 2 K \sqrt{d_2 T_m \log(T_m)} + \sqrt{\frac{8 K T}{\rho}} \\
        &\leq 2 K \sqrt{d_2 T_m \log(T_m)} + \sqrt{\frac{32 K T}{\rho}}
    \end{align*}
    
\end{proof}

%% file: regret_hylinucb_appendix.tex
\section{Regret Analysis of \HyLinUCB}
\label{appendix:hylinucb-regret-analysis}
In this section, we will use the same notations as specified in Appendix~\ref{appendix:proof-of-linucb} except, in this section $$\gamma = 2(\sqrt{2\lambda} S + \sqrt{2(d_1 + d_2) \log \frac{T}{\delta}})~.$$

\subsection{Proof of Theorem~\ref{theorem:regret-bound-hylinucb}}
\label{appendix:proof-of-hylinucb}

\begin{lemma}
\label{lemma:ucb-bonus-of-hylinucb}
Under event $\cE$ (defined in Appendix~\ref{appendix:proof-of-linucb}), for $t > T_o$, for all $i \in [K]$, $\lvert \dotp{\Tilde{x}_{i,t}}{\thtill_{t-1} - \thbar} \rvert \leq 2(\sqrt{2\lambda} S + \sqrt{2(d_1 + d_2) \log \frac{T}{\delta}})  \norm{\Tilde{x}_{i, t}}{\Mb_{t-1}^{-1}} + O(\frac{1}{\sqrt{t-1}})~.$
\end{lemma}

\begin{proof}
    Let us denote by $\Delta \thtill_{t-1} = \thtill_{t-1} - \thbar$. We can write,
    \begin{align*}
        \Ub_{t-1} \Delta \thtill_{t-1} &= \Mb_{t-1} \Delta \thtill_{t-1} + (\Ub_{t-1} - \Mb_{t-1})\Delta \thtill_{t-1} \\
        &= \Mb_{t-1} \Delta \thtill_{t-1} - \Ub_{t-1}^{1/2} \Ab_{t-1} \Ub_{t-1}^{1/2} \Delta \thtill_{t-1} \tag{by Proposition~\ref{proposition:relation-between-M-and-U-matrices}}
    \end{align*}
    Thus, $\Delta \thtill_{t-1} = \Ub_{t-1}^{-1} \Mb_{t-1} \Delta\thtill_{t-1} - \Ub_{t-1}^{-1/2} \Ab_{t-1} \Ub_{t-1}^{1/2} \Delta \thtill_{t-1}$. Consequently,
    \begin{align}
        \dotp{\Tilde{x}_{i, t}}{\Delta\thtill_{t-1}} &= \dotp{\Tilde{x}_{i, t}}{\Ub_{t-1}^{-1} \Mb_{t-1}\Delta\thtill_{t-1}} - \dotp{\Tilde{x}_{i, t}}{\Ub_{t-1}^{-1/2} \Ab_{t-1} \Ub_{t-1}^{1/2} \Delta \thtill_{t-1}} \nonumber \\
        &= \dotp{\Tilde{x}_{i, t}}{\Ub_{t-1}^{-1} \left( \sum_{s=1}^{t-1} \eta_s \Tilde{x}_s - \lambda \thbar \right)} - \dotp{\Tilde{x}_{i, t}}{\Ub_{t-1}^{-1/2} \Ab_{t-1} \Ub_{t-1}^{1/2} \Delta \thtill_{t-1}} \tag{by Proposition~\ref{proposition:expression-for-M-(thtill-thbar)}} \nonumber \\
        &= \dotp{\Ub_{t-1}^{-1} \Tilde{x}_{i, t}}{\sum_{s=1}^{t-1} \eta_s \Tilde{x}_s - \lambda \thbar} -  \dotp{\Tilde{x}_{i, t}}{\Ub_{t-1}^{-1/2} \Ab_{t-1} \Ub_{t-1}^{1/2} \Delta \thtill_{t-1}} \label{eq:tilde-x-dot-product-with-del-phi}
    \end{align}
    Since $\Ub_{t-1}$ is block diagonal, $\Ub^{-1}_{t-1} = \diag(\Vb_{t-1}^{-1}, \Wb_{1,t-1}^{-1}, \dots, \Wb_{K,t-1}^{-1})$. Also, $\tilde{x}_{i,t}$ is a sparse vector, with zeros everywhere except first $d_1$ coordinates and coordinates starting from $d_1 + (i - 1) d_2 + 1$ till $d_1 + i d_2$. Thus, $\Ub_{t-1}^{-1} \tilde{x}_{i,t} = \begin{bmatrix}
        (\Vb_{t-1}^{-1} x_{i,t})^\intercal & \zero & \dots & (\Wb_{i, t-1}^{-1} z_{i,t})^\intercal & \dots & \zero
    \end{bmatrix}^\intercal$. Therefore, 
    \begin{align*}
    \dotp{\Ub_{t-1}^{-1} \tilde{x}_{i,t}}{\sum_{s=1}^{t-1} \eta_s \Tilde{x}_s - \lambda \thbar} &= \dotp{\Vb_{t-1}^{-1} x_{i,t}}{\sum_{s=1}^{t-1} \eta_s x_s - \lambda \thta} \\
    & \quad\quad+ \dotp{\Wb_{i, t-1}^{-1} z_{i, t}}{\sum_{s\in \cT_{i, t-1}}\eta_s z_s - \lambda \bta{i}}~.
    \end{align*}
    Plugging this in eq.~\eqref{eq:tilde-x-dot-product-with-del-phi},
    \begin{align*}
        & \abr{\dotp{\Tilde{x}_{i, t}}{\Delta\thtill_{t-1}}} \\ 
        &= \left\lvert \dotp{\Vb_{t-1}^{-1} x_{i,t}}{\sum_{s=1}^{t-1} \eta_s x_s - \lambda \thta} + \dotp{\Wb_{i, t-1}^{-1} z_{i, t}}{\sum_{s\in \cT_{i, t-1}} z_s - \lambda \bta{i}} \right. \\
        &\quad\quad\quad \left. -  \dotp{\Tilde{x}_{i, t}}{\Ub_{t-1}^{-1/2} \Ab_{t-1} \Ub_{t-1}^{1/2} \Delta \thtill_{t-1}} \right\rvert \\
        &\leq \lvert \dotp{\Vb_{t-1}^{-1} x_{i,t}}{\sum_{s=1}^{t-1} \eta_s x_s - \lambda \thta} \rvert + \lvert \dotp{\Wb_{i,t-1}^{-1} z_{i, t}}{\sum_{s\in \cT_{i, t-1}} z_s - \lambda \bta{i}} \rvert \\
        &\quad\quad\quad + \lvert \dotp{\Tilde{x}_{i, t}}{\Ub_{t-1}^{-1/2} \Ab_{t-1} \Ub_{t-1}^{1/2} \Delta \thtill_{t-1}} \rvert \\
        &\leq \norm{x_{i,t}}{\Vb_{t-1}^{-1}}\cdot \norm{\sum_{s=1}^{t-1} \eta_s x_s - \lambda \thta}{\Vb_{t-1}} + \norm{z_{i,t}}{\Wb_{i,t-1}^{-1}}\cdot \norm{\sum_{s \in \cT_{i,t-1}} \eta_s z_s - \lambda \bta{i}}{\Wb_{i,t-1}} \\
        &\quad\quad\quad + \lvert \dotp{\Tilde{x}_{i, t}}{\Ub_{t-1}^{-1/2} \Ab_{t-1} \Ub_{t-1}^{1/2} \Delta \thtill_{t-1}} \rvert \\
        &\leq (\sqrt{2d_1 \log(T/\delta)} + \sqrt{\lambda} S)  \norm{x_{i,t}}{\Vb_{t-1}^{-1}} + (\sqrt{2d_2 \log(\tau_{i,t-1}/\delta)} + \sqrt{\lambda} S)  \norm{z_{i, t}}{\Wb_{i,t-1}^{-1}}\\
        & \quad\quad\quad + \lvert \dotp{\Tilde{x}_{i, t}}{\Ub_{t-1}^{-1/2} \Ab_{t-1} \Ub_{t-1}^{1/2} \Delta \thtill_{t-1}} \rvert \tag{by Lemma~\ref{lemma:self-normalized-martingale-bound}}
    \end{align*}
    Next, we upper bound $\lvert \dotp{\Tilde{x}_{i, t}}{\Ub_{t-1}^{-1/2} \Ab_{t-1} \Ub_{t-1}^{1/2} \Delta \thtill_{t-1}} \rvert$ as follows:
    \begin{align*}
        \lvert \dotp{\Tilde{x}_{i, t}}{\Ub_{t-1}^{-1/2} \Ab_{t-1} \Ub_{t-1}^{1/2} \Delta \thtill_{t-1}} \rvert &\leq \lambda_{max}(\Ub_{t-1}^{-1/2} \Ab_{t-1} \Ub_{t-1}^{1/2}) \norm{\Tilde{x}_{i, t}}{2} \norm{\Delta \thtill_{t-1}}{2} \tag{Sub-multiplicativity of norm} \\
        &\leq \lambda_{max}(\Ab_{t-1})\cdot S \cdot \norm{\Delta \thtill_{t-1}}{2} \tag{Lemma~\ref{lemma:A-inv-B-A-has-same-norm-as-B}; Assumption~\ref{assumption:boundedness}}
    \end{align*}
    From proof of Lemma~\ref{lemma:norm-bound-of-Z}, it can be extracted that, 
    \begin{align*}
        \lambda_{max}(\Ab_{t})
        \leq 4K\sqrt{\log(K(d_1 + d_2)/\delta)}\rbr{ \frac{\sqrt{T_m}}{\sqrt{\rho \lambda t}} + \frac{\sqrt{2}}{\sqrt{\rho^2 t}} } 
    \end{align*}
    Next, we have that $\norm{\Delta \thtill_{t-1}}{2} \leq \frac{\norm{\Delta \thtill_{t-1}}{\Mb_{t-1}}}{\sqrt{\lambda}}$ since $\lambda_{min}(\Mb_{t-1}) \geq \lambda$. Lastly, via Lemma~\ref{lemma:theta-tilde-theta-bar-under-M-norm-confidence-set-guarantee}, $\norm{\Delta \thtill_{t-1}}{\Mb_{t-1}} \leq 2 S \sqrt{\lambda K} + \sqrt{2(d_1 + d_2 K) \log(\frac{T}{\delta})}$. Putting everything back together,
    \begin{align*}
       & \lvert \dotp{\Tilde{x}_{i, t}}{\Ub_{t-1}^{-1/2} \Ab_{t-1} \Ub_{t-1}^{1/2} \Delta \thtill_{t-1}} \rvert \\
       &\leq 4 K S \sqrt{\log(\frac{K(d_1 + d_2)}{\delta})} \rbr{ \frac{\sqrt{T_m}}{\sqrt{\rho \lambda t}} + \frac{\sqrt{2}}{\sqrt{\rho^2 t}} } \frac{2 S \sqrt{\lambda K} + \sqrt{2(d_1 + d_2 K) \log(\frac{T}{\delta})}}{\sqrt{\lambda}} \\
       &= \frac{\alpha(\lambda)}{\sqrt{t}} \tag{definition of $\alpha(\lambda)$}
    \end{align*}
    Thus, we have,
    \begin{align*}
        &\abr{\dotp{\Tilde{x}_{i, t}}{\Delta\thtill_{t-1}}}
        \leq \rbr{\sqrt{2d_1 \log(\frac{T}{\delta})}  + \sqrt{\lambda} S}  \norm{x_{i,t}}{\Vb_{t-1}^{-1}} \\
        &\quad\quad\quad\quad\quad\quad\quad\quad\quad + \rbr{\sqrt{2d_2 \log(\frac{\tau_{i,t-1}}{\delta})} + \sqrt{\lambda} S}  \norm{z_{i, t}}{\Wb_{i,t-1}^{-1}} + \frac{\alpha(\lambda)}{\sqrt{t-1}} \\
        &\leq \sqrt{2 \rbr{2 (d_1 + d_2) \log(\frac{T}{\delta}) + 2 \lambda S^2}\cdot \rbr{\norm{x_{i,t}}{\Vb_{t-1}^{-1}}^2 + \norm{z_{i, t}}{\Wb_{i,t-1}^{-1}}^2}} + \frac{\alpha(\lambda)}{\sqrt{t-1}} \tag{Cauchy-Schwarz} \\
        &= \sqrt{2 \rbr{2 (d_1 + d_2) \log(\frac{T}{\delta}) + 2 \lambda S^2}\cdot \norm{\tilde{x}_{i,t}}{\Ub_{t-1}^{-1}}^2} + \frac{\alpha(\lambda)}{\sqrt{t-1}} \\
        &\leq \sqrt{2 \rbr{2 (d_1 + d_2) \log(\frac{T}{\delta}) + 2 \lambda S^2}\cdot 2\norm{\tilde{x}_{i,t}}{\Mb_{t-1}^{-1}}^2}  + \frac{\alpha(\lambda)}{\sqrt{t-1}} \tag{Lemma~\ref{lemma:lowener-order-of-M-and-U}} \\
        &\leq 2 \rbr{\sqrt{2 (d_1 + d_2) \log(\frac{T}{\delta})} + S\sqrt{2 \lambda}} \norm{\tilde{x}_{i,t}}{\Mb_{t-1}^{-1}} + \frac{\alpha(\lambda)}{\sqrt{t-1}}
    \end{align*}
\end{proof}

\begin{lemma}
\label{lemma:instantaneous-regret-bound-of-hylinucb}
    Under event $\cE$, after $T_o$ rounds have elapsed, the instantaneous regret of \HyLinUCB is upper bounded as: $$\dotp{\tilde{x}_{i^*_t, t}}{\thbar} - \dotp{\tilde{x}_{i_t, t}}{\thbar} \leq 2 \gamma \norm{\tilde{x}_{i_t, t}}{\Mb_{t-1}^{-1}} + \frac{2\alpha(\lambda)}{\sqrt{t-1}}$$
\end{lemma}

\begin{proof}
    We have the following:
    \begin{align}
        \dotp{\tilde{x}_{i^*_t, t}}{\thbar} &\leq \dotp{\tilde{x}_{i^*_t, t}}{\thtill_{t-1}} + \gamma \norm{\tilde{x}_{i^*_t, t}}{\Mb_{t-1}^{-1}} + \frac{\alpha(\lambda)}{\sqrt{t-1}} \tag{Lemma~\ref{lemma:ucb-bonus-of-hylinucb}} \nonumber\\
        &\leq \dotp{\tilde{x}_{i_t, t}}{\thtill_{t-1}} + \gamma \norm{\tilde{x}_{i_t, t}}{\Mb_{t-1}^{-1}} + \frac{\alpha(\lambda)}{\sqrt{t-1}} \label{eq:upper-bound-of-optimal-arm-hylinucb}
    \end{align}
    where the second inequality is because $i_t = \argmax_{i\in[K]} \dotp{\tilde{x}_{i, t}}{\thtill_{t-1}} + \gamma \norm{\tilde{x}_{i, t}}{\Mb_{t-1}^{-1}}$. Similarly, we have:
    \begin{align}
        \dotp{\tilde{x}_{i_t, t}}{\thbar} &\geq \dotp{\tilde{x}_{i_t, t}}{\thtill_{t-1}} - \gamma \norm{\tilde{x}_{i_t, t}}{\Mb_{t-1}^{-1}} - \frac{\alpha(\lambda)}{\sqrt{t-1}} \label{eq:lower-bound-of-played-arm-hylinucb}
    \end{align}
    Using eq.~\ref{eq:upper-bound-of-optimal-arm-hylinucb} and~\ref{eq:lower-bound-of-played-arm-hylinucb},
    $    \dotp{\tilde{x}_{i^*_t, t}}{\thbar} - \dotp{\tilde{x}_{i_t, t}}{\thbar}
        \leq 2 \gamma \norm{\tilde{x}_{i_t, t}}{\Mb_{t-1}^{-1}} + \frac{2\alpha(\lambda)}{\sqrt{t-1}}$
\end{proof}

Now we restate Theorem~\ref{theorem:regret-bound-hylinucb} and present its proof.

\begin{theorem}[Regret of \HyLinUCB]
\label{theorem:regret-bound-hylinucb_appendix}
    At the end of $T$ rounds, the regret of \HyLinUCB\ (Algorithm~\ref{algo:hylinUCB} with $\lambda=K$, $\gamma = \sqrt{KS} + \sqrt{2(d_1 + d_2)\log(T/\delta)}$) under Assumptions \ref{assumption:independent-subgaussian-features}, \ref{assumption:boundedness} is upper bounded by 
    \[
     C_1 \sqrt{K^3 T \log({K(d_1 + d_2)}/{\delta})} + C_2 \sqrt{(d_1 + d_2 K) K T \log({K T (d_1 + d_2)}/{\delta})}
    \]
with probability at least $1 - 4\delta$. Here $C_1, C_2 > 0$ are universal constants and $T$ is assumed to be $\tilde{\Omega}(K^4)$.
\end{theorem}

\begin{proof}
    The proof is similar to the proof of Theorem~\ref{theorem:regret-bound-linucb_appendix}. We will assume throughout that event $\cE$ holds, which happens with probability at least $1 - 4 \delta$ (Lemma~\ref{lemma:condensed-lemma-for-linucb-proof}). Let $i^*_t$ and $i_t$ be the best arm and the arm played in round $t$ respectively. Then, following similar arguments as in the proof of Theorem~\ref{theorem:regret-bound-linucb_appendix}, we have,
    \begin{align*}
        &\Reg(T, \HyLinUCB) = \sum_{t=1}^T \dotp{\tilde{x}_{i^*_t, t}}{\thbar} - \dotp{\tilde{x}_{i_t, t}}{\thbar} \\
        &= \sum_{t=1}^{T_o} \dotp{\tilde{x}_{i^*_t, t}}{\thbar} - \dotp{\tilde{x}_{i_t, t}}{\thbar} + \sum_{t=T_o + 1}^T \dotp{\tilde{x}_{i^*_t, t}}{\thbar} - \dotp{\tilde{x}_{i_t, t}}{\thbar} \\
        &\leq 2 T_o S + \sum_{t=T_o}^T \dotp{\tilde{x}_{i^*_t, t}}{\thbar} - \dotp{\tilde{x}_{i_t, t}}{\thbar} \tag{rewards are bounded in $[-S,S]$} \\
        &\leq 2T_o S + 2 \gamma \sum_{t=T_o + 1}^T \norm{\tilde{x}_{i_t, t}}{\Mb_{t-1}^{-1}} + \sum_{t=T_o + 1}^{T} \frac{2 \alpha(\lambda)}{\sqrt{t-1}}  \tag{Lemma~\ref{lemma:instantaneous-regret-bound-of-hylinucb}} \\
        &\leq 2T_o S + 2\sqrt{2} \gamma \sum_{t=T_o + 1}^T \norm{\tilde{x}_{i_t, t}}{\Ub_{t-1}^{-1}} + \sum_{t=T_o + 1}^{T} \frac{2 \alpha(\lambda)}{\sqrt{t-1}} \tag{Lemma~\ref{lemma:lowener-order-of-M-and-U}} \\
        &= 2T_o S + \sqrt{8} \gamma \sum_{t=T_o + 1}^T \rbr{\norm{x_{i_t, t}}{\Vb_{t-1}^{-1}} + \norm{z_{i_t, t}}{\Wb_{i_t, t-1}^{-1}}} + \sum_{t=T_o + 1}^{T} \frac{2 \alpha(\lambda)}{\sqrt{t-1}} \\
        &\leq 2T_o S + \sqrt{8} \gamma \rbr{2 K \sqrt{d_2 T_m \log(T_m)} + \sqrt{\frac{32 K T}{\rho}}} + \sum_{t=T_o + 1}^{T} \frac{2 \alpha(\lambda)}{\sqrt{t-1}} \tag{Lemma~\ref{lemma:cumulative-regret-bound-after-T_o-rounds}} \\
        &\leq 2T_o S + \sqrt{8} \gamma \rbr{2 K \sqrt{d_2 T_m \log(T_m)} + \sqrt{\frac{32 K T}{\rho}}} + \int_{t=1}^{T+1} \frac{2 \alpha(\lambda)}{\sqrt{t-1}} \\
        &\leq 2T_o S + \sqrt{8} \gamma \rbr{2 K \sqrt{d_2 T_m \log(T_m)} + \sqrt{\frac{32 K T}{\rho}}} + 4 \alpha(\lambda) \sqrt{T}
    \end{align*}
    Finally, from definition of $\alpha(\lambda)$, one can observe that setting $\lambda  = K$, and letting $c = 4S \sqrt{\log(K(d_1 + d_2)/\delta)}$ and $\beta = 2S \sqrt{\lambda K} + \sqrt{2(d_1 + d_2K) \log(T/\delta)}$
    \begin{align*}
        \alpha(\lambda) &= c \beta K \rbr{ \frac{\sqrt{T_m}}{\lambda \sqrt{\rho}} + \frac{\sqrt{2}}{\rho} \frac{1}{\sqrt{\lambda}}} \\
        &= c \rbr{\beta \sqrt{\frac{T_m}{\rho}} + \frac{\sqrt{2}}{\rho} \rbr{2S K^{3/2} + \sqrt{2(d_1 + d_2 K) K \log(T/\delta)}}} \\
        &\leq \frac{C_o}{\rho} \rbr{S^2 K^{3/2} +  S \sqrt{(d_1 + d_2 K) K \log(T/\delta)}} \sqrt{ \log(K(d_1 + d_2)/\delta)}
    \end{align*}
    for some universal constant $C_o$. Therefore, keeping only asymptotically dominating terms in $T$ and $K$,
    \begin{align*}
        \Reg(T, \HyLinUCB) &\leq C_1 \sqrt{K^3 T \log(\frac{K(d_1 + d_2)}{\delta})} \\
        &\quad\quad+ C_2 \sqrt{(d_1 + d_2 K) K T \log(\frac{K T (d_1 + d_2)}{\delta})}
    \end{align*}
    where $C_1, C_2 > 0$ are universal constants.
\end{proof}

%% file: regret_dislinucb_appendix.tex
\section{Regret Analysis of \DisLinUCB}
\label{appendix:dislinucb-regret-analysis}

In the algorithm of \DisLinUCB, since the parameters for each arm are estimated in a disjoint fashion, let us denote (with a slight abuse of notation) by $\phist{i} \coloneqq \begin{bmatrix} {\thta}^\intercal & {\bta{i}}^\intercal \end{bmatrix}^\intercal$ the true parameters that determine the reward of every arm $i \in [K]$. Moreover, we overload the notation $\thtill$ to denote by $\phiht{i} \coloneqq \begin{bmatrix}{\thth}^\intercal_i & {\bth{i}}^\intercal \end{bmatrix}^\intercal$ the estimate of $\phist{i}$ for every arm $i \in [K]$. For every round $t \in [T]$ and every arm $i \in [K]$, let $\overline{x}_{i,t} \coloneqq \begin{bmatrix} x_{i,t}^\intercal & z_{i,t}^\intercal\end{bmatrix}^\intercal$ be the feature vector of arm $i$ in round $t$ formed by appending the \emph{shared} and the \emph{action} features. Finally, we define $\Mb_{i,t} \coloneqq \sum_{s \in \cT_{i,t}} \overline{x}_{i,s} \overline{x}_{i,s}^\intercal + \lambda \Ib$ for this section. Since \DisLinUCB also solves a ridge regression to obtain the parameter estimates, we have, after round $t$,
\begin{align}
    \phiht{i,t} = \Mb_{i,t}^{-1} \sum_{s \in \cT_{i,t}} r_s \overline{x}_{s} , \quad \forall\ i \in [K]
\end{align}

\subsection{Proof of Corollary~\ref{corollary:regret-bound-DisLinUCB} from Section~\ref{subsection:linucb-dislinucb}}
\label{subsection:regret-analysis-of-DisLinUCB-main-proof}

First, we will present the key lemmas that will be used to prove Corollary~\ref{corollary:regret-bound-DisLinUCB}. Then the detailed proof is presented which follows the same recipe used in proving Theorem~\ref{theorem:regret-bound-linucb_appendix}.

\begin{proposition}
    For all $i \in [K]$ and $t \geq 0$, $\Mb_{i,t} \rbr{\phiht{i,t} - \phist{i}} = \sum_{s \in \cT_{i,t}} \eta_s \overline{x}_s - \lambda \phist{i}$
\end{proposition}

\begin{lemma}
\label{lemma:DisLinUCB-confidence-ellipsoid}
    With probability at least $1 - \delta$, for all $i \in [K]$ and $t \geq 0$, we have that $\norm{\phiht{i,t} - \phist{i}}{\Mb_{i,t}} \leq \gamma \coloneqq 2\sqrt{\lambda S} + \sqrt{2(d_1 + d_2) \log(KT/\delta)}$
\end{lemma}

\begin{proof}
    The proof is a direct application of Lemma~\ref{lemma:abbasi-yadkori-confidence-bound-of-estimator} with the fact that $\norm{\phist{i}}{2} = \sqrt{\norm{\thta}{2}^2 + \norm{\bta{i}}{2}^2} \leq \sqrt{2 S^2} = \sqrt{2}S \leq 2 \sqrt{S}$.
\end{proof}

\begin{lemma}
\label{lemma:DisLinUCB-estimation-error-at-x_i-guarantee}
    For any given $i \in [K]$ and $t \geq 0$, with probability at least $1 - \delta$, we have $\abr{\dotp{\overline{x}_{i,t}}{\phiht{i,t} - \phist{i}}} \leq \gamma \norm{\overline{x}_{i,t}}{\Mb_{i,t}^{-1}}$
\end{lemma}

\begin{proof}
    $\lvert \dotp{\Tilde{x}}{\thtill_t - \thbar}\rvert \leq \norm{\Tilde{x}}{\Mb_t^{-1}} \cdot \norm{\thtill_t - \thbar}{\Mb_t}$ by Cauchy-Schwarz inequality. Thereby, applying Lemma~\ref{lemma:theta-tilde-theta-bar-under-M-norm-confidence-set-guarantee} finishes the proof.
\end{proof}

Recall that $T_m = \rbr{\frac{16}{\rho^2} + \frac{8}{3\rho}} \log\rbr{\frac{2(d_1 + d_2) K T}{\delta}}$ and define $$T'_o \coloneqq \big(\frac{128}{\rho^2} \log(K(d_1 + d_2) / \delta)\big) \vee T_m~.$$

Next, we define the following events which will be used in the regret analysis of \DisLinUCB:

\begin{align*}
    \cE_1 &= \cbr{\forall\ i \in [K], \forall\ t \geq 0, \text{ s.t. } \tau_{i,t} \geq T'_o: \lambda_{min}\rbr{\Mb_{i,t}} \geq \rho \tau_{i,t} / 4} \\
    \cE_2 &= \cbr{\forall\ i \in [K], \forall\ t \geq 0, \norm{\phiht{i,t} - \phist{i}}{\Mb_{i,t}} \leq \gamma} \\
    \cE &= \cE_1 \cap \cE_2
\end{align*}

\begin{lemma} 
\label{lemma:good-event-probability-for-DisLinUCB}
    $\PP[\cE] \geq 1 - 4 \delta$
\end{lemma}

\begin{proof}
    By Corollary~\ref{corollary:min-eigenvalue-of-M_i,t-DisLinUCB}, $\PP[\cE_1] \geq 1 - 3\delta$ and by Lemma~\ref{lemma:DisLinUCB-confidence-ellipsoid}, $\PP[\cE_2] \geq 1 - \delta$. Thus, by Union Bound, $\PP[\cE] \geq 1 - 4 \delta$.
\end{proof}

Now, we restate Corollary~\ref{corollary:regret-bound-DisLinUCB} and present its proof.

\begin{corollary}[Regret of \DisLinUCB]
\label{corollary:regret-bound-DisLinUCB_appendix}
At the end of $T$ rounds, the regret of \DisLinUCB\ (Algorithm \ref{algo:dislinUCB} with $\lambda=1$, $\gamma = 2\sqrt{S} + \sqrt{2(d_1 + d_2) \log(KT/\delta)}$) under Assumptions \ref{assumption:independent-subgaussian-features}, \ref{assumption:boundedness} is upper bounded by $C \sqrt{(d_1 + d_2) K T \log(K T / \delta)}$ with probability at least $1 - 4\delta$. Here, $C > 0$ is a universal constant and $T$ is assumed to be $\tilde{\Omega}(1)$. 
\end{corollary}

\begin{proof}
    Throughout the proof, we will assume that event $\cE$ holds, which happens with probability at least $1 - 4 \delta$. Let $i^*_t$ denote the best arm in round $t$. Further, to simplify notation, let $\phiht{i,t-1}$ be written as $\phiht{i}$ for all $i \in [K]$. The instantaneous regret in round $t$ is upper bounded as:
    \begin{align*}
        &\dotp{\overline{x}_{i^*_t, t}}{\phist{i^*_t}} - \dotp{\overline{x}_{i_t, t}}{\phist{i_t}} \\
        &\leq \dotp{\overline{x}_{i^*_t, t}}{\phiht{i^*_t}} + \gamma \norm{\overline{x}_{i^*_t, t}}{\Mb^{-1}_{i^*_t, t-1}} - \dotp{\overline{x}_{i_t, t}}{\phiht{i_t}} + \gamma \norm{\overline{x}_{i_t, t}}{\Mb^{-1}_{i_t, t-1}} \tag{Lemma~\ref{lemma:DisLinUCB-estimation-error-at-x_i-guarantee}}\\
        &\leq \dotp{\overline{x}_{i_t, t}}{\phiht{i_t}} + \gamma \norm{\overline{x}_{i_t, t}}{\Mb^{-1}_{i_t, t-1}} - \dotp{\overline{x}_{i_t, t}}{\phiht{i_t}} + \gamma \norm{\overline{x}_{i_t, t}}{\Mb^{-1}_{i_t, t-1}} \tag{$i_t = \argmax_{i \in [K]} \dotp{\overline{x}_{i, t}}{\phiht{i}} + \gamma \norm{\overline{x}_{i, t}}{\Mb^{-1}_{i, t-1}}$} \\
        &= 2 \gamma \norm{\overline{x}_{i_t, t}}{\Mb^{-1}_{i_t, t-1}}
    \end{align*}
    Thus, the cumulative regret of \DisLinUCB over $T$ rounds can be upper bounded by:
    \begin{align}
    \label{eq:DisLinUCB-regret-upper-bound-in-terms-of-UCB-bonus}
        \Reg(T, \DisLinUCB) \leq \sum_{t = 1}^T 2 \gamma \norm{\overline{x}_{i_t, t}}{\Mb^{-1}_{i_t, t-1}} = 2 \gamma \sum_{i = 1}^K \sum_{s \in \cT_{i,T}} \norm{\overline{x}_{i, s}}{\Mb^{-1}_{i, s-1}}
    \end{align}
    Now, for a given $i \in [K]$, let the elements in $\cT_{i,T}$ be rearranged so that they are in increasing order. Let $\cT_{i,\xi}$ denote the subset of $\cT_{i,T}$ that contains only the first $T'_o$ elements if $\abr{\cT_{i,T}} > T'_o$ else $\cT_{i,\xi} = \cT_{i,T}$. Finally, let $\tau_{i,\xi} = \abr{\cT_{i,\xi}}$. Thus, we can write the sum in the RHS above as:
    \begin{align*}
        \sum_{i = 1}^K \sum_{s \in \cT_{i,T}} \norm{\overline{x}_{i, s}}{\Mb^{-1}_{i, s-1}} &= \sum_{i=1}^K \sum_{s \in \cT_{i, \xi}} \norm{\overline{x}_{i, s}}{\Mb^{-1}_{i, s-1}} + \sum_{i: \tau_{i,T} > T_o} \sum_{s \in \cT_{i,T} \setminus \cT_{i,\xi}} \norm{\overline{x}_{i, s}}{\Mb^{-1}_{i, s-1}}
    \end{align*}
    To upper bound the first term above, we will use Elliptic Potential Lemma (Lemma~\ref{lemma:elliptic-potential-lemma}) but for the second term above, since the matrices $\Mb_{i,s}$ contains the sum of at least $T_o$ rank one (of the type $\overline{x} \overline{x}^\intercal$) terms, we can apply the eigenvalue guarantee of event $\cE$. Specifically,
    \begin{align*}
        &\sum_{i=1}^K \sum_{s \in \cT_{i, \xi}} \norm{\overline{x}_{i, s}}{\Mb^{-1}_{i, s-1}} + \sum_{i: \tau_{i,T} > T'_o} \sum_{s \in \cT_{i,T} \setminus \cT_{i,\xi}} \norm{\overline{x}_{i, s}}{\Mb^{-1}_{i, s-1}} \\
        &\leq \sum_{i=1}^K \sqrt{2 (d_1 + d_2) \abr{\cT_{i, \xi}} \log\rbr{\abr{\cT_{i, \xi}}}} + \sum_{i: \tau_{i,T} > T'_o} \sum_{s \in \cT_{i,T} \setminus \cT_{i,\xi}} \norm{\overline{x}_{i, s}}{\Mb^{-1}_{i, s-1}} \tag{Lemma~\ref{lemma:elliptic-potential-lemma} on first term} \\
        &\leq \sum_{i=1}^K \sqrt{2 (d_1 + d_2) T'_o \log\rbr{T'_o}} + \sum_{i: \tau_{i,T} > T'_o} \sum_{s \in \cT_{i,T} \setminus \cT_{i,\xi}} \norm{\overline{x}_{i, s}}{\Mb^{-1}_{i, s-1}} \tag{$\abr{\cT_{i, \xi}} < T'_o\ \forall\ i \in [K]$} \\
        &\leq K\sqrt{2 (d_1 + d_2) T'_o \log\rbr{T'_o}} + \sum_{i: \tau_{i,T} > T'_o} \sum_{s = T'_o + 1}^{\tau_{i, T}} \frac{2\sqrt{2}}{\sqrt{\rho (s-1)}} \tag{$\lambda_{min}\rbr{\Mb_{i, s}} \geq \rho \tau_{i,s} / 4$ under $\cE$; $\norm{\overline{x}_{i,s}}{2} \leq \sqrt{2}$} \\
        &\leq K\sqrt{2 (d_1 + d_2) T'_o \log\rbr{T'_o}} + \sum_{i: \tau_{i,T} > T'_o} \int_{s = 1}^{\tau_{i, T} + 1} \frac{2\sqrt{2}}{\sqrt{\rho (s-1)}} \\
        &= K\sqrt{2 (d_1 + d_2) T'_o \log\rbr{T'_o}} + \sum_{i: \tau_{i,T} > T'_o} \frac{4\sqrt{2}}{\sqrt{\rho}} \sqrt{\tau_{i,T}} \\
        &\leq K\sqrt{2 (d_1 + d_2) T'_o \log\rbr{T'_o}} + \frac{4\sqrt{2}}{\sqrt{\rho}} \sum_{i = 1}^K \sqrt{\tau_{i, T}} \\
        &\leq K\sqrt{2 (d_1 + d_2) T'_o \log\rbr{T'_o}} + \frac{4\sqrt{2}}{\sqrt{\rho}} \sqrt{K \sum_{i=1}^K \tau_{i,T}} \tag{Cauchy-Schwarz} \\
        &= K\sqrt{2 (d_1 + d_2) T'_o \log\rbr{T'_o}} + \frac{4\sqrt{2}}{\sqrt{\rho}} \sqrt{K T}
    \end{align*}
    Thus, by plugging this upper bound back in eq.~\eqref{eq:DisLinUCB-regret-upper-bound-in-terms-of-UCB-bonus}, we obtain,
    \begin{align*}
        \Reg(T, \DisLinUCB) \leq 2 \gamma \rbr{K\sqrt{2 (d_1 + d_2) T'_o \log\rbr{T'_o}} + \frac{4\sqrt{2}}{\sqrt{\rho}} \sqrt{K T}}
    \end{align*}
    Noting that $\gamma = O(\sqrt{(d_1 + d_2)}\log(K T/\delta))$ and writing only the asymptotically dominating term, we have, for a universal constant $C > 0$,
    \begin{align*}
        \Reg(T, \DisLinUCB) \leq C \sqrt{(d_1 + d_2) K T \log(K T / \delta)}
    \end{align*}
\end{proof}

\input{dislinucb_supp_lemmas_appendix}

%% file: dislinucb_supp_lemmas_appendix.tex
\subsection{Supporting Lemmas for Appendix~\ref{subsection:regret-analysis-of-DisLinUCB-main-proof}}
\label{appendix:supporting-lemmas-for-DisLinUCB-analysis}

Let us define $\Ub_{i,t} = \begin{bmatrix}
    \Vb_{i,t} & \Bb_{i,t} \\
    \Bb^\intercal_{i,t} & \Wb_{i,t}
\end{bmatrix}$ where $\Vb_{i,t} = \lambda \Ib + \sum_{s \in \cT_{i,t}} x_{i,s} x_{i,s}^\intercal$, $\Wb_{i,t} = \lambda \Ib + \sum_{s \in \cT_{i,t}} z_{i,s} z_{i,s}^\intercal$ and $\Bb_{i,t} = \sum_{s \in \cT_{i,t}} x_{i,s} z_{i,s}^\intercal$ for all $i \in [K]$.

\begin{proposition}
\label{proposition:relation-between-M-and-U-in-DisLinUCB}
    For any $i \in [K]$ and $t \geq 0$,
    \begin{align*}
    \Ub_{i,t}^{-\frac{1}{2}} \Mb_{i,t} \Ub_{i,t}^{-\frac{1}{2}} = \Ib + \begin{bmatrix}
        \zero & \Vb_{i,t}^{-\frac{1}{2}} \Bb_{i,t} \Wb_{i,t}^{-\frac{1}{2}} \\
        \Wb_{i,t}^{-\frac{1}{2}} \Bb_{i,t}^\intercal \Vb_{i,t}^{-\frac{1}{2}} & \zero
    \end{bmatrix} \equiv \Ib + \Ab_{i,t} \tag{definition of $\Ab_{i,t}$}
    \end{align*}
\end{proposition}

\begin{proof}
    The proof is by directly substituting $\Ub_{i,t}$ and $\Mb_{i,t}$ and noting that since $\Ub_{i,t}$ is a block diagonal matrix, $\Ub_{i,t}^{-\frac{1}{2}} = \diag\rbr{\Vb_{i,t}^{-\frac{1}{2}}, \Wb_{i,t}^{-\frac{1}{2}}}$.
\end{proof}

\begin{lemma}
    Let $\Zb_{i,t} = \Vb_{i,t}^{-\frac{1}{2}} \Bb_{i,t} \Wb_{i,t}^{-\frac{1}{2}}$ for all $i \in [K]$ and $t \geq 0$. Then, almost surely, $\lambda_{max}(\Ab_{i,t}) = -\lambda_{min}(\Ab_{i,t}) = \sigma_{max}(\Zb_{i,t})$.
\end{lemma}

\begin{proof}
    The proof follows the same set of arguments presented in the proof of Lemma~\ref{lemma:eigenvalue-properties-of-A-in-terms-of-Z}. Specifically, since $\Ab_{i,t} = \begin{bmatrix}
        \zero & \Zb_{i,t} \\
        \Zb_{i,t}^\intercal & \zero
    \end{bmatrix}$ is the Hermitian dilation of $\Zb_{i,t}$, thus by Lemma~\ref{lemma:hermitian-dilation-eigenvalue-relation}, the square of eigenvalues of $\Ab_{i,t}$ are equal to the square of the singular values of $\Zb_{i,t}$. Consequently, for every positive singular value $\sigma$ of $\Zb_{i,t}$, there are two eigenvalues of $\Ab_{i,t}$ namely, $\sigma$ and $-\sigma$. Hence, clearly, $\lambda_{max}(\Ab_{i,t}) = \sigma_{max}(\Zb_{i,t})$ and $\lambda_{min}(\Ab_{i,t}) = - \sigma_{max}(\Zb_{i,t})$.
\end{proof}
Now, we will turn to upper bounding $\norm{\Zb_{i,t}}{}$ for all $i \in [K]$.

Hereon, let us define the event $$\cE_1 \coloneqq \left\{\forall\ i \in [K], t \geq 0,\  \norm{\Bb_{i,t}}{} \leq \sqrt{8 \tau_{i,t} \log\rbr{\frac{K (d_1 + d_2)}{\delta}}}\right\}$$ Thus, from Corollary~\ref{corollary:concentration-of-B-matrix-for-all-i-and-all-t} in Appendix~\ref{appendix:supporting-lemmas-for-LinUCB-analysis}, we have that $\PP[\cE_1] \geq 1 - \delta$.

\begin{lemma}
\label{lemma:min-eigenvalue-under-stochasticity-problem-based-result-DisLinUCB}
    We first define the quantity $T_m \coloneqq \rbr{\frac{16}{\rho^2} + \frac{8}{3\rho}} \log\rbr{\frac{2(d_1 + d_2) K T}{\delta}}$. Then, $\PP\sbr{\forall i \in [K], t \geq 0 \text{ s.t. } \tau_{i,t} \geq T_m, \lambda_{min}(\Vb_{i,t}) \geq \frac{\rho}{2}\tau_{i,t}} \geq 1 -\delta$, and $\PP\big[\forall i \in [K], t \geq 0 \text{ s.t. }  \tau_{i,t} \geq T_m,\ \lambda_{min}(\Wb_{i,t}) \geq \frac{\rho}{2}\tau_{i,t} \big] \geq 1 -\delta$.
\end{lemma}

\begin{proof}
    The result is obtained by directly applying Lemma~\ref{lemma:min-eigenvalue-bartlett} followed by union bound.
\end{proof}

\begin{align}\cE_2 &= \{ \forall i \in [K], t \geq 0 \text{ s.t. } \tau_{i,t} \geq T_m, \lambda_{min}(\Vb_{i,t}) \geq \frac{\rho}{2}\tau_{i,t} \} \nonumber \\ 
\cE_3 &= \{\forall i \in [K], t \geq 0 \text{ s.t. }  \tau_{i,t} \geq T_m,\ \lambda_{min}(\Wb_{i,t}) \geq \frac{\rho}{2}\tau_{i,t}\} \nonumber \\
\cE_o &\coloneqq \cE_1 \cap \cE_2 \cap \cE_3 \label{eq:definition-of-good-event-E_o}
\end{align}

\begin{corollary}
\label{corollary:concentration-good-event-probability-dislinucb}
    $\PP[\cE_o] \geq 1 - 3\delta~.$
\end{corollary}

\begin{proof}
    By Union Bound, $\PP[\overline{\cE_o}] \leq \PP[\overline{\cE_1}] + \PP[\overline{\cE_2}] + \PP[\overline{\cE_3}] \leq 3 \delta~.$
\end{proof}

\begin{lemma}
    Under event $\cE_o$, if for a given $i \in [K]$, $\tau_{i,t} \geq T_o \coloneqq \big(\frac{128}{\rho^2} \log(K(d_1 + d_2) / \delta)\big) \vee T_m$, then $\sigma_{max}\rbr{\Zb_{i,t}} \leq \frac{1}{2}$.
\end{lemma}

\begin{proof}
    \begin{align*}
        \norm{\Zb_{i,t}}{} &= \norm{\Vb_{i,t}^{-\frac{1}{2}} \Bb_{i,t} \Wb_{i,t}^{-\frac{1}{2}}}{} \\
        &\leq \norm{\Vb_{i,t}^{-\frac{1}{2}}}{} \norm{\Bb_{i,t}}{} \norm{\Wb_{i,t}^{-\frac{1}{2}}}{} \tag{Sub-multiplicativity of norm} \\
        &= \frac{\norm{\Bb_{i,t}}{}}{\sqrt{\lambda_{min}\rbr{\Vb_{i,t}}} \sqrt{\lambda_{min}\rbr{\Wb_{i,t}}}} \\
        &\leq \frac{\sqrt{8 \tau_{i,t} \log(K(d_1 + d_2)/\delta)}}{\sqrt{\rho \tau_{i,t}/2} \sqrt{\rho \tau_{i,t}/2}} \tag{Event $\cE_o$ holds, $\tau_{i,t} \geq T_m$} \\
        &= \sqrt{\frac{32 \log(K(d_1 + d_2)/\delta)}{\rho^2 \tau_{i,t}}} \\
        &\leq \frac{1}{2} \tag{$\tau_{i,t} \geq 128 \log(K(d_1 + d_2) / \delta) / \rho^2$}
    \end{align*}
\end{proof}

\begin{lemma}
\label{lemma:lowener-order-of-M-and-U-DisLinUCB}
    Under event $\cE_o$, for all $i \in [K]$ such that $\tau_{i,t} \geq T_o$, $\frac{1}{2}\Ub_{i,t} \preccurlyeq \Mb_{i,t} \preccurlyeq 2\Ub_{i,t}$.
\end{lemma}

\begin{proof}
    We first note that by Lemma~\ref{lemma:eigenvalue-properties-of-A-in-terms-of-Z}, $\lambda_{max}(\Ab_{i,t}) = - \lambda_{min}(\Ab_{i,t})$. Thus, under $\cE_o$ and $\tau_{i,t} \geq T_o$, $\lambda_{max}(\Ab_{i,t}) = \norm{\Zb_{i,t}}{} \leq \frac{1}{2}$ implies that $\lambda_{min}(\Ab_{i,t}) \geq -\frac{1}{2}$. Thus, all eigenvalues of $\Ab_{i,t}$ lie in the range $[-1/2, 1/2]$. This in turn implies that the eigenvalues of $\Ib + \Ab_{i,t}$ lie in the range $[1/2, 3/2]$. From this, it can be concluded that, $$\frac{1}{2}\Ib \preccurlyeq \Ib + \Ab_{i,t} \preccurlyeq 2\Ib$$ as eigenvalues of $2\Ib$ dominate that of $\Ib + \Ab_{i,t}$ which in turn dominate that of $(1/2)\Ib$. Thus, we have, by combining with Proposition~\ref{proposition:relation-between-M-and-U-in-DisLinUCB}, $\frac{1}{2}\Ib \preccurlyeq \Ub_{i,t}^{-\frac{1}{2}} \Mb_{i,t} \Ub_{i,t}^{-\frac{1}{2}} \preccurlyeq 2\Ib$, which then implies that,
    \begin{align*}
        \frac{1}{2}\Ub_{i,t} \preccurlyeq \Mb_{i,t} \preccurlyeq 2\Ub_{i,t}
    \end{align*}
\end{proof}

\begin{corollary}
\label{corollary:min-eigenvalue-of-M_i,t-DisLinUCB}
    $\PP\sbr{\forall i \in [K], t \geq 0 \text{ s.t. } \tau_{i,t} \geq T_o: \lambda_{min}(\Mb_{i,t}) \geq \rho \tau_{i,t} / 4} \geq 1 - 3 \delta$.
\end{corollary}

\begin{proof}
    Assume that event $\cE_o$ holds, which happens with probability $1 - 3 \delta$ (Corollary~\ref{corollary:concentration-good-event-probability-dislinucb}). Then, by Lemma~\ref{lemma:lowener-order-of-M-and-U-DisLinUCB}, $\Mb_{i,t} \succcurlyeq \frac{1}{2}\Ub_{i,t}$. Thus, $\lambda_{min}(\Mb_{i,t}) \geq \lambda_{min}(\frac{1}{2}\Ub_{i,t}) = \frac{1}{2} \lambda_{min}(\diag(\Vb_{i,t}, \Wb_{i,t})) = \frac{1}{2} (\lambda_{min}(\Vb_{i,t}) \wedge \lambda_{min}(\Wb_{i,t})) \geq  \frac{\rho \tau_{i,t}}{4}$.
\end{proof}

%% file: concentration_appendix.tex
\section{Concentration Results of Random Matrices and Vectors}
\label{appendix:concentration-results}
\begin{lemma}[Lemma 7 of~\citep{chatterji2020osom}]
\label{lemma:min-eigenvalue-bartlett} Suppose $\{x_s\}_{s=1}^t$ is a stochastic process in $\RR^d$ such that for the filtration $\cF_t = \{x_s\}_{s=1}^{t}$, it holds that $\EE\sbr{x_t \mid \cF_{t-1}} = \zero$ and $\EE\sbr{x_t x_t^\intercal \mid \cF_{t-1}} \succcurlyeq \rho \Ib$. Moreover, suppose $\norm{x_s}{2} \leq 1$ for all $s \geq 1$. Define the matrix $\Vb_t$ as
\begin{align*}
    \Vb_t := \Ib + \sum_{s=1}^{t} x_{s} x_{s}^{\intercal}.
\end{align*}
Then, with probability at least $1-\delta$, we have
\begin{align*}
\lambda_{min}(\Vb_t) \geq 1+ \frac{\rho t}{2}
\end{align*} for all $\left(16/\rho^2 + 8/(3\rho)\right)\log\left(2dT/\delta\right) \leq t \leq T$.
\end{lemma}

\begin{lemma}[Matrix Azuma (Theorem 7.1~\citep{tropp2012user}]
\label{lemma:matrix-azuma}
    Let $\{\Xb_k\}_{k=1}^\infty$ be a matrix martingale difference sequence in $\RR^{d_1 \times d_2}$ and let $\cH(\Xb_k)$ be its Hermitian Dilation (definition~\ref{definition-of-Hermitian-dilation}). Further, let $\{\Ab_k\}_{k=1}^\infty$ be a fixed sequence of matrices in $\RR^{(d_1 + d_2) \times (d_1 + d_2)}$ such that,
    \begin{align*}
        \EE_{k-1}[\Xb_k] = \zero \quad \text{and} \quad \cH(\Xb_k)^2 \preccurlyeq \Ab_k^2 \quad \text{almost surely.}
    \end{align*}
    Further, let $\sigma^2_t = \lambda_{max}\rbr{\sum_{k=1}^t \Ab_k^2}$ for $t \geq 1$. Then, for all $\varepsilon \geq 0$, $$\PP\sbr{\exists t \geq 1: \sigma_{max}\rbr{\sum_{k=1}^t \Xb_k} \geq \varepsilon} \leq (d_1 + d_2)\exp\rbr{-\frac{\varepsilon^2}{8 \sigma^2_t}}$$
\end{lemma}

\begin{lemma}[Theorem 1 of~\cite{abbasi2011improved}]
\label{lemma:self-normalized-martingale-bound}
    Let $\{\cF_t\}_{t=1}^\infty$ be a filtration. Let $\{x_t\}_{t=1}^\infty$ be a stochastic process in $\RR^d$ with $\norm{x_t}{2} \leq 1$ such that $x_t$ is $\cF_t$ measurable. Let $\{\eta_t\}_{t=1}^\infty$ be a real-valued stochastic process such that $\eta_t$ is $\cF_t$-measurable and $\eta_t$ is conditionally $1$-sub-Gaussian, i.e., $\EE[\exp(\ell\eta_t)\mid \cF_t] \leq \exp(\ell^2/2)$ for all $\ell \in \RR$. Let $\lambda > 0$ and for any $t \geq 1$ define:
    \begin{align*}
        S_t = \sum_{s=1}^{t} \eta_s x_s \quad\quad\quad \Hb_t = \sum_{s=1}^{t} x_s x_s^\intercal + \lambda \Ib
    \end{align*}
    Then, for any $\delta \in (0,1)$, with probability at least $1-\delta$, for all $0 \leq t \leq T$
    \begin{align*}
        \norm{S_t}{\Hb^{-1}_t} \leq \sqrt{d \log\left(\frac{1 + t/d\lambda}{\delta}\right)} \leq \sqrt{2d\log(T/\delta)}
    \end{align*}
    
\end{lemma}

\begin{lemma}[Theorem 2 of~\cite{abbasi2011improved}]
\label{lemma:abbasi-yadkori-confidence-bound-of-estimator}
    Let $\thth$ be the least squares estimate $\Hb_t^{-1} \left(\sum_{s=1}^t y_s x_s\right)$ over the feature-target pairs $\{(x_s, y_s)\}_{s=1}^t$, where $y_s = \dotp{x_s}{\thta} + \eta_s$, and $\eta_s$ and $\Hb_t$ is as defined in Lemma~\ref{lemma:self-normalized-martingale-bound}. Let $\norm{\thta}{} \leq S$ and $\norm{x_s}{} \leq 1$ for all $s \in [t]$. Then, for all $0 \leq t \leq T$, with probability at least $1 - \delta$,
    \begin{align*}
        \norm{\thth - \thta}{\Hb_t} \leq \sqrt{\lambda}S + \sqrt{2d\log(T/\delta)}
    \end{align*}
\end{lemma}






%% file: lin_alg_appendix.tex
\section{Some Useful Results}
\label{section:linear-algebra-results}

\begin{lemma}[{~\citep[Section 0.8.5]{horn2012matrix}}]
\label{lemma:block-matrix-determinant}
Let $\Mb = \begin{bmatrix}
    \Ab & \Bb \\ \Cb & \Db
\end{bmatrix}$ be a block matrix with invertible $\Ab$ and $\Db$. Then, $$\det(\Mb) = \det(\Ab)\det(\Db - \Cb \Ab^{-1} \Bb)$$
\end{lemma}

\begin{definition}[Hermitian Dilation]
\label{definition-of-Hermitian-dilation}
    The Hermitian dilation of a rectangular matrix $\Bb \in \RR^{d_1 \times d_2}$ is defined as the Hermitian matrix $$\Hb = \begin{bmatrix}
    \zero & \Bb \\
    \Bb^\intercal & \zero
\end{bmatrix}$$
Note that $\Hb^2 = \begin{bmatrix}
    \Bb \Bb^\intercal & \zero \\
    \zero & \Bb^\intercal \Bb
\end{bmatrix}~.$
\end{definition}

\begin{lemma}[{\citep[Eq. (2.1.28)]{tropp15matrix}}]
\label{lemma:hermitian-dilation-eigenvalue-relation}
The squared singular values of a rectangular matrix $\Bb$ coincides with the squared eigenvalues of its Hermitian dilation $\Hb$.
\end{lemma}

\begin{lemma}[Elliptic Potential Lemma {~\citep[Lemma 10]{abbasi2011improved}}]\label{lemma:elliptic-potential-lemma}
    Let $x_1, x_2, \dots x_t$ be a sequence of vectors in ${\RR}^d$ and let $\norm{x_s}{2} \leq 1$ for all $s \in [t]$. Further, let $\Vb_s = \sum_{m=1}^{s-1} x_m x_m^\intercal + \lambda \Ib$. Suppose $\lambda \geq 1$. Then,
    \begin{align}
        \sum_{s=1}^t \norm{x_s}{\Vb_s^{-1}}^2 \leq 2d \log\left(1 + \frac{t}{\lambda d}\right) \leq 4d \log(t)
    \end{align}
\end{lemma}

\begin{lemma}[{\citep[Theorem 1.3.3]{horn2012matrix}}]
\label{lemma:A-inv-B-A-has-same-norm-as-B}
    Suppose $\Ab$ is an invertible and $\Xb$ is a symmetric matrix. Then, eigenvalues of $\Ab^{-1} \Xb \Ab$ are the same as eigenvalues of $\Xb$. Particularly, $\lambda_{max}\rbr{\Ab^{-1} \Xb \Ab} = \lambda_{max}\rbr{\Xb}$.
\end{lemma}



%% file: empirical-validation-assumption.tex
\section{Empirical Validation of Assumption~\ref{assumption:independent-subgaussian-features}}
\label{appendix:empirical-validation}
Although the previous works make this assumption for the completely shared setting, which is indeed algorithm dependent, there are problem instances in the shared setting that allow all algorithms to satisfy this assumption (see~\citep[Section 3]{pmlr-v139-papini21a}). However, to verify whether the diversity assumption is a valid assumption in the hybrid setting, we perform experiments with algorithms \LinUCB, \HyLinUCB and \DisLinUCB. The key implication of these assumptions on which the improved regret guarantees of \HyLinUCB and \LinUCB rely is that the minimum eigenvalue of the following matrices should grow linearly: (i) $\Vb_t = \Ib + \sum_{s=1}^t x_{i_s,s} x_{i_s,s}^\intercal$ and (ii) $\Wb_{i,t} = \Ib + \sum_{s=1}^t z_{i_s,s} z_{i_s, s}^\intercal \ind{i_s = i}$. Besides, the maximum singular value of $\Bb_{i,t} = \sum_{s=1}^t x_{i_s,s} z_{i_s, s}^\intercal \ind{i_s = i}$ for all $i \in [K]$ should grow as $\sqrt{\tau_{i,t}}$ (the number of times arm $i$ is pulled till time $t$) for \HyLinUCB and \LinUCB. Lastly, for \DisLinUCB, this assumption's key implication is that the minimum eigenvalue of $\Ib + \sum_{s=1}^t \overline{x}_{i_s,s} \overline{x}_{i_s,s}^\intercal$ grows linearly with $\tau_{i,t}$, where $\overline{x}_{i_s,s} = \begin{bmatrix} x_{i_s,s}^\intercal & z_{i_s,s}^\intercal \end{bmatrix}^\intercal$.

\begin{figure}[bt]
    \centering
    \begin{subfigure}{0.6 \textwidth}
        \includegraphics[width=\textwidth]{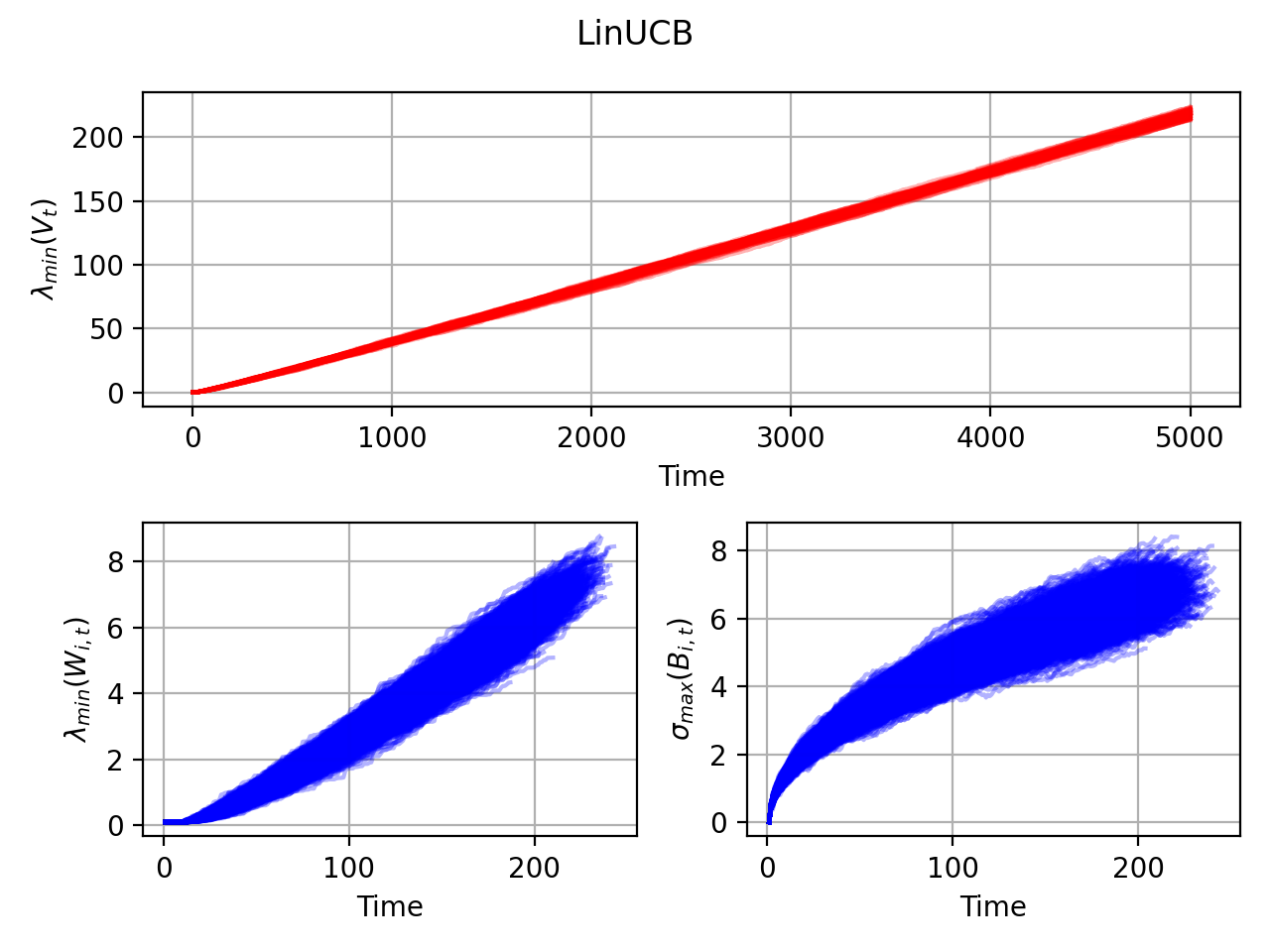}
    \end{subfigure} \\
    \begin{subfigure}{0.6 \textwidth}
        \includegraphics[width=\textwidth]{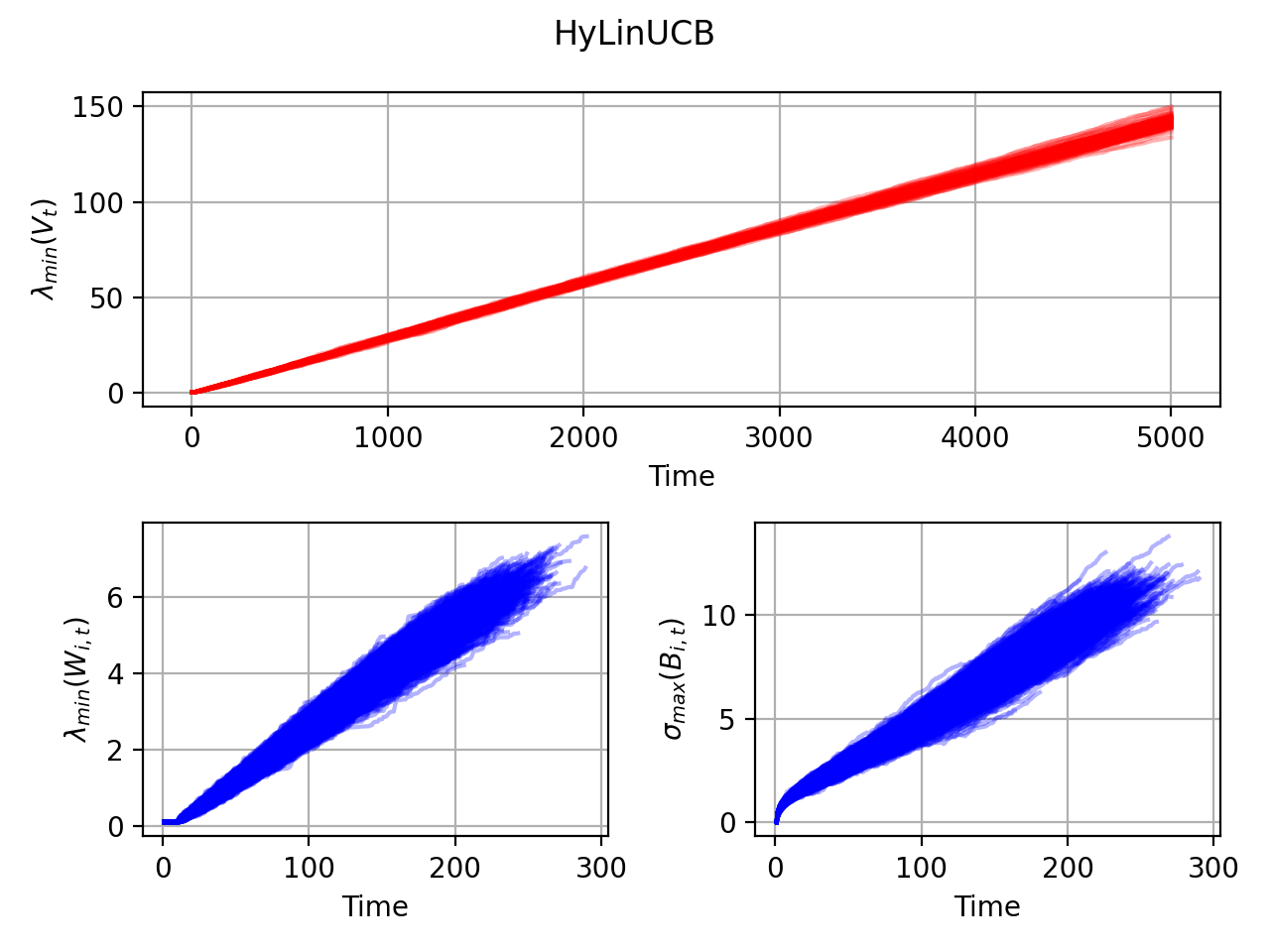}
    \end{subfigure} \\
    \begin{subfigure}{0.4 \textwidth}
        \includegraphics[width=\textwidth]{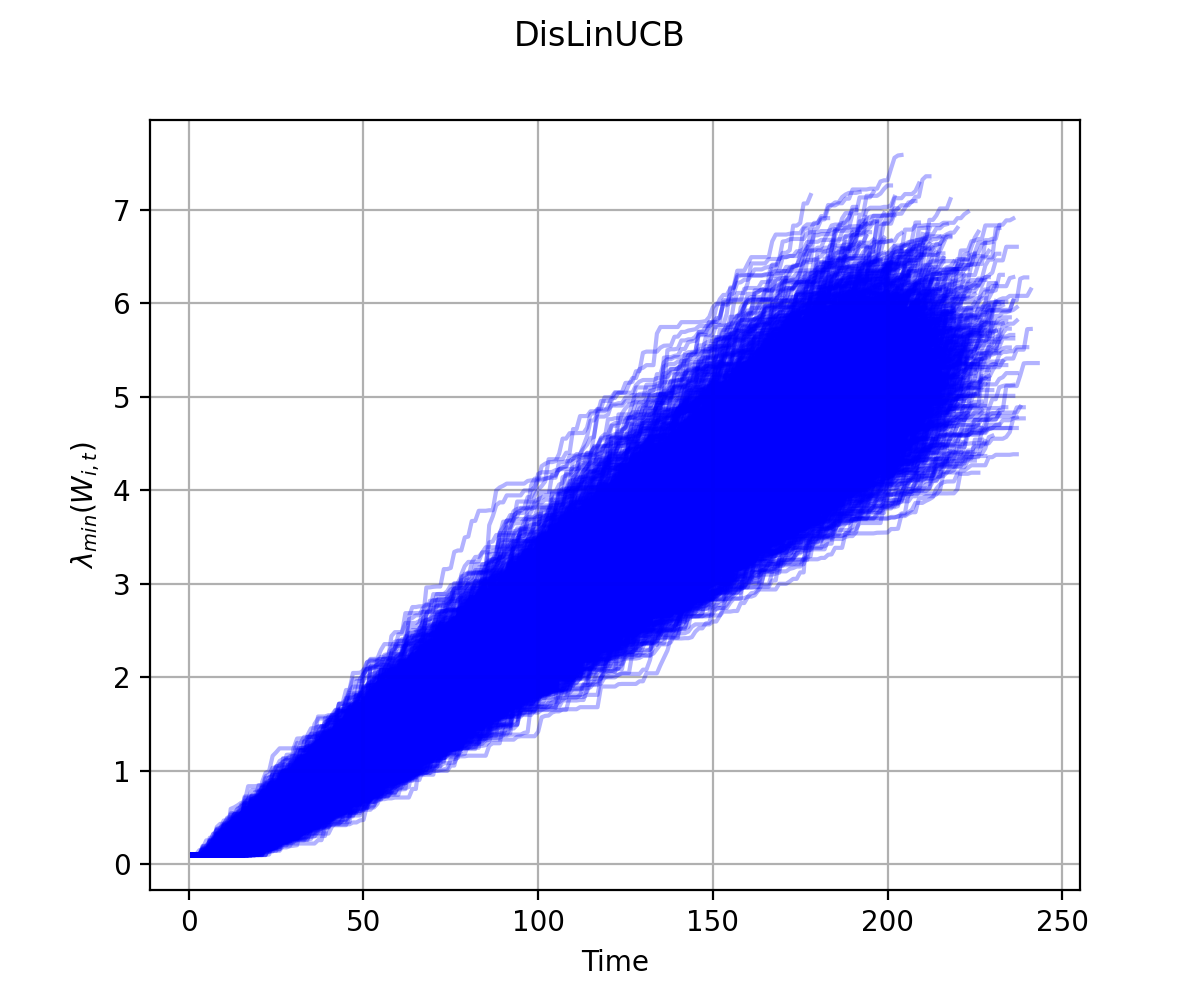}
    \end{subfigure}
    \caption{\textbf{Empirical validation of the key implications of Assumption~\ref{assumption:independent-subgaussian-features}}. From top to bottom: \LinUCB, \HyLinUCB and \DisLinUCB. In \LinUCB: top plot shows minimum eigenvalue of $\Vb_t$; bottom left plot shows minimum eigenvalue of $\Wb_{i,t}$; bottom right shows maximum singular value of $\Bb_{i,t}$. Similar scheme is followed for \HyLinUCB. In \DisLinUCB, only the maximum eigenvalue of the relevant matrix is shown.}
    \label{fig:assumption-figure}
\end{figure}

Setting $d_1 = d_2 = 10$, $K = 25$ and $T = 5000$, we randomly sample a realization of the reward parameters $\thta$ and $\{\bta{i}\}_{i \in [K]}$ uniformly from a unit ball of appropriate dimensions, and then perform $N = 100$ trials of the algorithms. The arm features are also generated by uniformly sampling from the respective unit balls in every round in every trial. The results are presented in Fig.~\ref{fig:assumption-figure}. From the result, it can be noted that (i) the minimum eigenvalue of the relevant matrices grows linearly with rounds (or number of arm pulls) for all the algorithms and (ii) the maximum singular value of the relevant matrices grow as $\sqrt{t}$ with round $t$ in both \LinUCB and \HyLinUCB. These are the chief implications of Assumption~\ref{assumption:independent-subgaussian-features} used in the proofs of the regret guarantees. The experimental validation therefore provides strong grounds for the diversity assumption.

%% file: arxiv_main.bbl
\begin{thebibliography}{25}
\providecommand{\natexlab}[1]{#1}
\providecommand{\url}[1]{\texttt{#1}}
\expandafter\ifx\csname urlstyle\endcsname\relax
  \providecommand{\doi}[1]{doi: #1}\else
  \providecommand{\doi}{doi: \begingroup \urlstyle{rm}\Url}\fi

\bibitem[Abbasi-Yadkori et~al.(2011)Abbasi-Yadkori, P{\'a}l, and
  Szepesv{\'a}ri]{abbasi2011improved}
Yasin Abbasi-Yadkori, D{\'a}vid P{\'a}l, and Csaba Szepesv{\'a}ri.
\newblock Improved algorithms for linear stochastic bandits.
\newblock \emph{Advances in neural information processing systems}, 24, 2011.

\bibitem[Abeille and Lazaric(2017)]{Abeille17}
Marc Abeille and Alessandro Lazaric.
\newblock {Linear Thompson Sampling Revisited}.
\newblock In \emph{Proceedings of the 20th International Conference on
  Artificial Intelligence and Statistics}, volume~54 of \emph{Proceedings of
  Machine Learning Research}, pages 176--184. PMLR, 2017.

\bibitem[Agrawal and Goyal(2013)]{Agarwal2013}
Shipra Agrawal and Navin Goyal.
\newblock Thompson sampling for contextual bandits with linear payoffs.
\newblock In \emph{Proceedings of the 30th International Conference on
  International Conference on Machine Learning - Volume 28}, ICML'13, page
  III–1220–III–1228. JMLR.org, 2013.

\bibitem[Auer(2003)]{auer2003}
Peter Auer.
\newblock Using confidence bounds for exploitation-exploration trade-offs.
\newblock \emph{J. Mach. Learn. Res.}, 3\penalty0 (null):\penalty0 397–422,
  mar 2003.
\newblock ISSN 1532-4435.

\bibitem[Chatterji et~al.(2020)Chatterji, Muthukumar, and
  Bartlett]{chatterji2020osom}
Niladri Chatterji, Vidya Muthukumar, and Peter Bartlett.
\newblock Osom: A simultaneously optimal algorithm for multi-armed and linear
  contextual bandits.
\newblock In \emph{Proceedings of the Twenty Third International Conference on
  Artificial Intelligence and Statistics}, volume 108 of \emph{Proceedings of
  Machine Learning Research}, pages 1844--1854. PMLR, 26--28 Aug 2020.
\newblock URL \url{https://proceedings.mlr.press/v108/chatterji20b.html}.

\bibitem[Chu et~al.(2009)Chu, Park, Beaupre, Motgi, Phadke, Chakraborty, and
  Zachariah]{chu2009ACS}
Wei Chu, Seung-Taek Park, Todd Beaupre, Nitin Motgi, Amit Phadke, Seinjuti
  Chakraborty, and Joe Zachariah.
\newblock A case study of behavior-driven conjoint analysis on yahoo!: front
  page today module.
\newblock In \emph{Knowledge Discovery and Data Mining}, 2009.
\newblock URL \url{https://api.semanticscholar.org/CorpusID:18101293}.

\bibitem[Chu et~al.(2011)Chu, Li, Reyzin, and Schapire]{chu2011contextual}
Wei Chu, Lihong Li, Lev Reyzin, and Robert Schapire.
\newblock Contextual bandits with linear payoff functions.
\newblock In \emph{Proceedings of the Fourteenth International Conference on
  Artificial Intelligence and Statistics}, pages 208--214. JMLR Workshop and
  Conference Proceedings, 2011.

\bibitem[Dani et~al.(2008)Dani, Hayes, and Kakade]{Dani2008}
Varsha Dani, Thomas~P. Hayes, and Sham~M. Kakade.
\newblock Stochastic linear optimization under bandit feedback.
\newblock In \emph{Annual Conference Computational Learning Theory}, 2008.
\newblock URL \url{https://api.semanticscholar.org/CorpusID:9134969}.

\bibitem[Gentile et~al.(2017)Gentile, Li, Kar, Karatzoglou, Zappella, and
  Etrue]{gentile2017context}
Claudio Gentile, Shuai Li, Purushottam Kar, Alexandros Karatzoglou, Giovanni
  Zappella, and Evans Etrue.
\newblock On context-dependent clustering of bandits.
\newblock In \emph{International Conference on machine learning}, pages
  1253--1262. PMLR, 2017.

\bibitem[Ghosh et~al.(2021)Ghosh, Sankararaman, and Kannan]{ghosh2021problem}
Avishek Ghosh, Abishek Sankararaman, and Ramchandran Kannan.
\newblock Problem-complexity adaptive model selection for stochastic linear
  bandits.
\newblock In \emph{International Conference on Artificial Intelligence and
  Statistics}, pages 1396--1404. PMLR, 2021.

\bibitem[Hill et~al.(2017)Hill, Nassif, Liu, Iyer, and Vishwanathan]{Hill2017}
Daniel~N. Hill, Houssam Nassif, Yi~Liu, Anand Iyer, and S.V.N. Vishwanathan.
\newblock An efficient bandit algorithm for realtime multivariate optimization.
\newblock In \emph{Proceedings of the 23rd ACM SIGKDD International Conference
  on Knowledge Discovery and Data Mining}, KDD '17, page 1813–1821, New York,
  NY, USA, 2017. Association for Computing Machinery.
\newblock ISBN 9781450348874.
\newblock URL \url{https://doi.org/10.1145/3097983.3098184}.

\bibitem[Horn and Johnson(2012)]{horn2012matrix}
Roger~A Horn and Charles~R Johnson.
\newblock \emph{Matrix analysis}.
\newblock Cambridge university press, 2012.

\bibitem[Kim et~al.(2023)Kim, Paik, and Oh]{Kim23}
Wonyoung Kim, Myunghee~Cho Paik, and Min~Hwan Oh.
\newblock Squeeze all: Novel estimator and self-normalized bound for linear
  contextual bandits.
\newblock In \emph{International Conference on Artificial Intelligence and
  Statistics}, volume 206 of \emph{Proceedings of Machine Learning Research},
  pages 3098--3124. {PMLR}, 2023.
\newblock URL \url{https://proceedings.mlr.press/v206/kim23d.html}.

\bibitem[Lai and Robbins(1985)]{Lai1985}
T.L Lai and Herbert Robbins.
\newblock Asymptotically efficient adaptive allocation rules.
\newblock \emph{Advances in Applied Mathematics}, 6\penalty0 (1):\penalty0
  4--22, 1985.
\newblock ISSN 0196-8858.

\bibitem[Lattimore and Szepesv{\'a}ri(2020)]{lattimore2020bandit}
Tor Lattimore and Csaba Szepesv{\'a}ri.
\newblock \emph{Bandit algorithms}.
\newblock Cambridge University Press, 2020.

\bibitem[Li et~al.(2010)Li, Chu, Langford, and Schapire]{li2010contextual}
Lihong Li, Wei Chu, John Langford, and Robert~E Schapire.
\newblock A contextual-bandit approach to personalized news article
  recommendation.
\newblock In \emph{Proceedings of the 19th international conference on World
  wide web}, pages 661--670, 2010.

\bibitem[Li et~al.(2017)Li, Lu, and Zhou]{Li2017}
Lihong Li, Yu~Lu, and Dengyong Zhou.
\newblock Provably optimal algorithms for generalized linear contextual
  bandits.
\newblock In \emph{Proceedings of the 34th International Conference on Machine
  Learning - Volume 70}, ICML'17, page 2071–2080. JMLR.org, 2017.

\bibitem[Li et~al.(2019)Li, Wang, and Zhou]{li2019nearly}
Yingkai Li, Yining Wang, and Yuan Zhou.
\newblock Nearly minimax-optimal regret for linearly parameterized bandits.
\newblock In \emph{Conference on Learning Theory}, pages 2173--2174. PMLR,
  2019.

\bibitem[Papini et~al.(2021)Papini, Tirinzoni, Restelli, Lazaric, and
  Pirotta]{pmlr-v139-papini21a}
Matteo Papini, Andrea Tirinzoni, Marcello Restelli, Alessandro Lazaric, and
  Matteo Pirotta.
\newblock Leveraging good representations in linear contextual bandits.
\newblock In \emph{Proceedings of the 38th International Conference on Machine
  Learning}, volume 139 of \emph{Proceedings of Machine Learning Research},
  pages 8371--8380. PMLR, 18--24 Jul 2021.
\newblock URL \url{https://proceedings.mlr.press/v139/papini21a.html}.

\bibitem[Rusmevichientong and Tsitsiklis(2008)]{Rusmevichientong2008}
Paat Rusmevichientong and John~N. Tsitsiklis.
\newblock Linearly parameterized bandits.
\newblock \emph{Math. Oper. Res.}, 35:\penalty0 395--411, 2008.
\newblock URL \url{https://api.semanticscholar.org/CorpusID:3204347}.

\bibitem[Stanimirovi{\'c} et~al.(2019)Stanimirovi{\'c}, Katsikis, and
  Kolundzija]{Stanimirovi2019InversionAP}
Predrag~S. Stanimirovi{\'c}, Vasilios~N. Katsikis, and Dejan Kolundzija.
\newblock Inversion and pseudoinversion of block arrowhead matrices.
\newblock \emph{Appl. Math. Comput.}, 341:\penalty0 379--401, 2019.
\newblock URL \url{https://api.semanticscholar.org/CorpusID:52988431}.

\bibitem[Thompson(1933)]{Thompson1933}
William~R Thompson.
\newblock {On the Likelihood that One Unknown Probability Exceeds Another in
  View of the Evidence of Two Samples}.
\newblock \emph{Biometrika}, 25\penalty0 (3-4):\penalty0 285--294, 1933.
\newblock ISSN 0006-3444.
\newblock \doi{10.1093/biomet/25.3-4.285}.
\newblock URL \url{https://doi.org/10.1093/biomet/25.3-4.285}.

\bibitem[Tropp(2012)]{tropp2012user}
Joel~A Tropp.
\newblock User-friendly tail bounds for sums of random matrices.
\newblock \emph{Foundations of computational mathematics}, 12:\penalty0
  389--434, 2012.

\bibitem[Tropp(2015)]{tropp15matrix}
Joel~A. Tropp.
\newblock An introduction to matrix concentration inequalities.
\newblock \emph{Found. Trends Mach. Learn.}, 8\penalty0 (1–2):\penalty0
  1–230, may 2015.
\newblock ISSN 1935-8237.
\newblock URL \url{https://doi.org/10.1561/2200000048}.

\bibitem[{Yahoo! Webscope}(2010)]{yahoo-dataset}
{Yahoo! Webscope}.
\newblock {Yahoo! Front Page Today Module User Click Log Dataset,} version 1.0.
\newblock \url{ http://research.yahoo.com/Academic_Relations a}, 2010.
\newblock Accessed: 2024-01-22.

\end{thebibliography}
